\documentclass{article}

\PassOptionsToPackage{numbers, compress}{natbib}

\usepackage[preprint]{nips_2018}

\usepackage{array}
\usepackage{algorithm}
\usepackage{algpseudocode}
\usepackage{amsmath}
\usepackage{amssymb}
\usepackage{bbold}
\usepackage{bm}
\usepackage{colortbl}
\usepackage{graphicx}
\usepackage{multirow}
\usepackage{wrapfig}
\usepackage{subcaption}
\usepackage{tikz}
\usetikzlibrary{shapes, snakes, positioning, arrows,calc}
\newcommand{\mtt}[1]{\mathtt{#1}}
\newcolumntype{?}{!{\vrule width 1pt}}

\usepackage[compact]{titlesec}

\usepackage{titlesec}
\titlespacing\section{0pt}{2pt plus 2pt minus 2pt}{-2pt plus 2pt minus 0pt}
\titlespacing\subsection{0pt}{0pt plus 2pt minus 6pt}{-2pt plus 2pt minus 0pt}
\titlespacing\subsubsection{0pt}{2pt plus 2pt minus 2pt}{-2pt plus 2pt minus 0pt}
\titlespacing\paragraph{0pt}{2pt plus 2pt minus 2pt}{-2pt plus 2pt minus 0pt}

\usepackage{placeins}

\newcommand{\xhat}{\hat{x}}
\newcommand{\mbf}[1]{\mathbf{#1}}

\usepackage[utf8]{inputenc} 
\usepackage[T1]{fontenc}    
\usepackage{hyperref}       
\usepackage{url}            
\usepackage{booktabs}       
\usepackage{amsfonts}       
\usepackage{nicefrac}       
\usepackage{microtype}      

\title{{\Large A Unified View of Piecewise Linear Neural Network Verification}}

\author{Rudy Bunel\\
  University of Oxford\\
  \small{\texttt{rudy@robots.ox.ac.uk}}
  \And
  Ilker Turkaslan\\
  University of Oxford\\
  \small{\texttt{ilker.turkaslan@lmh.ox.ac.uk}}
  \And
  Philip H.S. Torr\\
  University of Oxford\\
  \small{\texttt{philip.torr@eng.ox.ac.uk}}
  \And
  Pushmeet Kohli\\
  Deepmind\\
  \small{\texttt{pushmeet@google.com}}
  \And
  M. Pawan Kumar\\
  University of Oxford \\
  Alan Turing Institute\\
  \small{\texttt{pawan@robots.ox.ac.uk}}
}

\begin{document}

\maketitle

\begin{abstract}
  The success of Deep Learning and its potential use in many safety-critical
  applications has motivated research on formal verification of Neural Network
  (NN) models. Despite the reputation of learned NN models to behave as black
  boxes and the theoretical hardness of proving their properties, researchers
  have been successful in verifying some classes of models by exploiting their
  piecewise linear structure and taking insights from formal methods such as
  Satisifiability Modulo Theory. These methods are however still far from
  scaling to realistic neural networks. To facilitate progress on this crucial
  area, we make two key contributions. First, we present a unified framework
  that encompasses previous methods. This analysis results in the identification
  of new methods that combine the strengths of multiple existing approaches,
  accomplishing a speedup of two orders of magnitude compared to the previous
  state of the art. Second, we propose a new data set of benchmarks which
  includes a collection of previously released testcases. We use the benchmark
  to provide the first experimental comparison of existing algorithms and
  identify the factors impacting the hardness of verification problems.
\end{abstract}

\section{Introduction}
Despite their success in a wide variety of applications, Deep Neural Networks
have seen limited adoption in safety-critical settings. The main explanation for
this lies in their reputation for being black-boxes whose behaviour can not be
predicted. Current approaches to evaluate trained models mostly rely on testing
using held-out data sets. However, as Edsger W. Dijkstra
said~\citep{buxton1970}, ``testing shows the presence, not the absence of
bugs''. If deep learning models are to be deployed to applications such as
autonomous driving cars, we need to be able to verify safety-critical
behaviours.


To this end, some researchers have tried to use formal methods. To the best of
our knowledge, \citet{Zakrzewski2001} was the first to propose a method to
verify simple, one hidden layer neural networks. However, only recently were
researchers able to work with non-trivial models by taking advantage of the
structure of ReLU-based networks~\citep{Cheng2017a,Katz2017}. Even then, these
works are not scalable to the large networks encountered in most real world
problems.

This paper advances the field of NN verification by making the following key
contributions:
\begin{enumerate}
\item We reframe state of the art verification methods as special cases of
  Branch-and-Bound optimization, which provides us with a \textbf{unified
    framework} to compare them.
\item We gather a data set of test cases based on the existing literature and
  extend it with new benchmarks. We \textbf{provide the first experimental
    comparison} of verification methods.
\item Based on this framework, we identify algorithmic improvements in the
  verification process, specifically in the way bounds are computed, the type of
  branching that are considered, as well as the strategies guiding the
  branching. Compared to the previous state of the art, these improvements lead
  to \textbf{speed-up of almost two orders of magnitudes}.
\end{enumerate}

Section 2 and 3 give the specification of the problem and formalise the
verification process. Section 4 presents our unified framework, showing that
previous methods are special cases and highlighting potential improvements.
Section 5 presents our experimental setup and Section 6 analyses the results.

\section{Problem specification}
We now specify the problem of formal verification of neural networks. Given a
network that implements a function $\mbf{\xhat_n} = f(\mbf{x_0})$, a bounded input domain
$\mathcal{C}$ and a property $P$, we want to prove
\begin{equation}
  \mbf{x_0} \in \mathcal{C},\quad \mbf{\xhat_n} = f(\mbf{x_0}) \implies P(\mbf{\xhat_n}).
  \label{eq:prob-form}
\end{equation}
For example, the property of robustness to adversarial examples in
$\mathcal{L}_{\infty}$ norm around a training sample $\mbf{a}$ with label $y_a$ would be encoded by
using \mbox{$\mathcal{C} \triangleq \left\{\mbf{x_0} | \ \|\mbf{x_0} -
      \mbf{a}\|_{\infty} \leq \epsilon\right\}$} and
  \mbox{$P(\mbf{\xhat_n}) =\left\{ \forall y \quad \xhat_{n
        [y_a]} \geq \xhat_{n [y]}\right\}$}.

  In this paper, we are going to focus on Piecewise-Linear Neural Networks
  (PL-NN), that is, networks for which we can decompose $\mathcal{C}$ into a set
  of polyhedra $\mathcal{C}_i$ such that \mbox{$\mathcal{C} = \cup_i\
    \mathcal{C}_i$}, and the restriction of $f$ to $\mathcal{C}_i$ is a linear
  function for each $i$. While this prevents us from including networks that use
  activation functions such as sigmoid or tanh, PL-NNs allow the use of linear
  transformations such as fully-connected or convolutional layers, pooling units
  such as MaxPooling and activation functions such as ReLUs. In other words,
  PL-NNs represent the majority of networks used in practice. Operations such as
  Batch-Normalization or Dropout also preserve piecewise linearity at
  test-time. 

The properties that we are going to consider are Boolean formulas over linear
inequalities. In our robustness to adversarial example above, the property is a
conjunction of linear inequalities, each of which constrains the output of the
original label to be greater than the output of another label.

The scope of this paper does not include approaches relying on additional
assumptions such as twice differentiability of the network
\citep{Hein2017, Zakrzewski2001}, limitation of the activation to binary values
\citep{Cheng2017a,Narodytska2017} or restriction to a single linear
domain~\citep{Bastani2016}. Since they do not provide formal guarantees, we also
don't include approximate approaches relying on a limited set of
perturbation~\citep{Huang2017} or on over-approximation methods that potentially
lead to undecidable properties~\citep{Pulina2010,Xiang2017}.

\section{Verification Formalism}
\subsection{Verification as a Satisfiability problem}
\label{subsec:verif-as-opt}
The methods we involve in our comparison all leverage the piecewise-linear
structure of PL-NN to make the problem more tractable. They all follow the same
general principle: given a property to prove, they attempt to discover a
counterexample that would make the property false. This is accomplished by
defining a set of variables corresponding to the inputs, hidden units and output
of the network, and the set of constraints that a counterexample would
satisfy.

To help design a unified framework, we reduce all instances of verification
problems to a canonical representation. Specifically, the whole satisfiability
problem will be transformed into a global optimization problem where the
decision will be obtained by checking the sign of the minimum.

If the property to verify is a simple inequality \mbox{$P(\mbf{\xhat_n})
  \triangleq \mbf{c}^T \mbf{\xhat_n} \geq b$}, it is sufficient to add to the
network a final fully connected layer with one output, with weight of $\mbf{c}$
and a bias of $-b$. If the global minimum of this network is positive, it
indicates that for all $\mbf{\xhat_n}$ the original network can output, we have
\mbox{$\mbf{c}^T \mbf{\xhat_n} -b \geq 0 \implies \mbf{c}^T \mbf{\xhat_n} \geq
  b$}, and as a consequence the property is True. On the other hand, if the
global minimum is negative, then the minimizer provides a counter-example. The
supplementary material shows that \texttt{OR} and \texttt{AND} clauses in the
property can similarly be expressed as additional layers, using MaxPooling units.

We can formulate any Boolean formula over linear inequalities on the output of
the network as a sequence of additional linear and max-pooling layers. The
verification problem will be reduced to the problem of finding whether the
scalar output of the potentially modified network can reach a negative value.
Assuming the network only contains ReLU activations between each layer, the
satisfiability problem to find a counterexample can be expressed as:
\begin{subequations}
  \begin{minipage}{.25\textwidth}
    \begin{align}
    &\mbf{l_0} \leq \mbf{x_0} \leq \mbf{u_0} \label{eq:ctx-bounds}\\
    &\xhat_{n} \leq 0 \label{eq:final}
    \end{align}
  \end{minipage}%
  \begin{minipage}{.75\textwidth}
    \begin{align}
      \qquad\mbf{\xhat_{i+1}} = W_{i+1} \mbf{x_i} + \mbf{b_{i+1}} \qquad&\forall i \in \{0, \ n-1\}\label{eq:lin-op}\\
      \qquad\mbf{x_i} = \max\left(\mbf{\xhat_i}, 0\right) \ \ \quad\qquad&\forall i \in \{1, \ n-1\} .\label{eq:relu-op}
    \end{align}
  \end{minipage}
  \label{eq:sat-problem}
\end{subequations}

Eq.~\eqref{eq:ctx-bounds} represents the constraints on the input and
Eq.~\eqref{eq:final} on the neural network output.
Eq.~\eqref{eq:lin-op} encodes the linear layers of the network and
Eq.~\eqref{eq:relu-op} the ReLU activation functions. If an assignment to all
the values can be found, this represents a counterexample. If this problem is
unsatisfiable, no counterexample can exist, implying that the property is True.
We emphasise that we are required to prove that no counter-examples can exist,
and not simply that none could be found.

While for clarity of explanation, we have limited ourselves to the specific case
where only ReLU activation functions are used, this is not restrictive. The
supplementary material contains a section detailing how each method specifically
handles MaxPooling units, as well as how to convert any MaxPooling operation into
a combination of linear layers and ReLU activation functions.

The problem described in~\eqref{eq:sat-problem} is still a hard problem. The
addition of the ReLU non-linearities~\eqref{eq:relu-op} transforms a problem
that would have been solvable by simple Linear Programming into an NP-hard
problem~\citep{Katz2017}. Converting a verification problem into this canonical
representation does not make its resolution simpler but it does provide a
formalism advantage. Specifically, it allows us to prove complex properties,
containing several \texttt{OR} clauses, with a single procedure rather than
having to decompose the desired property into separate queries as was done in
previous work~\citep{Katz2017}.

Operationally, a valid strategy is to impose the constraints
\mbox{\eqref{eq:ctx-bounds} to \eqref{eq:relu-op}} and minimise the value of
$\xhat_{n}$. Finding the exact global minimum is not necessary for verification.
However, it provides a measure of satisfiability or unsatisfiability. If the
value of the global minimum is positive, it will correspond to the margin by
which the property is satisfied.


\subsection{Mixed Integer Programming formulation}
\label{subsec:mip}
A possible way to eliminate the non-linearities is to encode them with the help
of binary variables, transforming the PL-NN verification
problem~\eqref{eq:sat-problem} into a Mixed Integer Linear Program (MIP). This
can be done with the use of ``big-M'' encoding. The following encoding is from
\citet{Tjeng2017}. Assuming we have access to lower and upper bounds on the
values that can be taken by the coordinates of $\mbf{\xhat_i}$, which we denote
$\mbf{l_i}$ and $\mbf{u_i}$, we can replace the non-linearities:
\begin{subequations}
  \begin{flalign}
    x_i = \max\left(\mbf{\xhat_i}, 0\right) \quad \Rightarrow \quad \bm{\delta_i} \in
    \{0,1\}^{h_i}, &\quad\mbf{x_i} \geq 0, \ \qquad  \mbf{x_i} \leq \mbf{u_i} \cdot \bm{\delta_i}\label{eq:mip_0}\\
    & \quad \mbf{x_i}\geq \mbf{\xhat_i},   \qquad \mbf{x_i} \leq \mbf{\xhat_i} - \mbf{l_i}\cdot(1 - \bm{\delta_i})\label{eq:mip_inp}
  \end{flalign}
  \label{eq:mip-form}%
\end{subequations}
It is easy to verify that \mbox{$\delta_{i[j]}=0 \Leftrightarrow x_{i[j]}=0$}
(replacing $\delta_{i[j]}$ in Eq.~\eqref{eq:mip_0}) and \mbox{$\delta_{i[j]}=1
  \Leftrightarrow x_{i[j]}=\xhat_{i[j]}$} (replacing $\delta_{i[j]}$ in
Eq.~\eqref{eq:mip_inp}).

By taking advantage of the feed-forward structure of the neural network, lower
and upper bounds $\mbf{l_i}$ and $\mbf{u_i}$ can be obtained by applying
interval arithmetic~\citep{Hickey2001} to propagate the bounds on the inputs, one
layer at a time.



Thanks to this specific feed-forward structure of the problem, the generic,
non-linear, non-convex problem has been rewritten into an MIP. Optimization of
MIP is well studied and highly efficient off-the-shelf solvers exist. As solving
them is NP-hard, performance is going to be dependent on the quality of both the
solver used and the encoding. We now ask the following question: how much
efficiency can be gained by using a bespoke solver rather than a generic one? In
order to answer this, we present specialised solvers for the PLNN verification
task.

\section{Branch-and-Bound for Verification}

As described in Section~\ref{subsec:verif-as-opt}, the verification problem can
be rephrased as a global optimization problem. Algorithms such as
Stochastic Gradient Descent are not appropriate as they have no way of
guaranteeing whether or not a minima is global. In this section, we present an
approach to estimate the global minimum, based on the Branch and Bound paradigm
and show that several published methods, introduced as examples of
Satisfiability Modulo Theories, fit this framework.

Algorithm~\ref{alg:bab} describes its generic form. The input domain is
repeatedly split into sub-domains (line 7), over which lower and upper bounds of
the minimum are computed (lines 9-10). The best upper-bound found so far serves
as a candidate for the global minimum. Any domain whose lower bound is greater
than the current global upper bound can be pruned away as it cannot contain the
global minimum (line 13, lines 15-17). By iteratively splitting the domains, it
is possible to compute tighter lower bounds. We keep track of the global lower
bound on the minimum by taking the minimum over the lower bounds of all
sub-domains (line 19). When the global upper bound and the global lower bound
differ by less than a small scalar $\epsilon$ (line 5), we consider that we have
converged.

\noindent Algorithm~\ref{alg:bab} shows how to optimise and obtain the global minimum.
If all that we are interested in is the satisfiability problem, the procedure
can be simplified by initialising the global upper bound with 0 (in line 2). Any
subdomain with a lower bound greater than 0 (and therefore not eligible to
contain a counterexample) will be pruned out (by line 15). The computation of
the lower bound can therefore be replaced by the feasibility problem (or its
relaxation) imposing the constraint that the output is below zero without
changing the algorithm. If it is feasible, there might still be a counterexample
and further branching is necessary. If it is infeasible, the subdomain can be
pruned out. In addition, if any upper bound improving on 0 is found on a
subdomain (line 11), it is possible to stop the algorithm as this already
indicates the presence of a counterexample.

\begin{wrapfigure}[22]{L}{.5\textwidth}
  \begin{minipage}{.5\textwidth}
    \vskip -20pt
    \begin{algorithm}[H]
      \caption{Branch and Bound}\label{alg:bab}
      \small
      \begin{algorithmic}[1]
        \algrenewcommand\algorithmicindent{1.0em}%
        \Function{BaB}{$\mtt{net}, \mtt{domain}, \epsilon$}
        \State $\mtt{global\_ub} \gets \inf$
        \State $\mtt{global\_lb} \gets - \inf$
        \State $\mtt{doms}\gets \left[ (\mtt{global\_lb}, \mtt{domain}) \right]$
        \While{$\mtt{global\_ub} - \mtt{global\_lb} > \epsilon $}
        \State $(\_\ , \mtt{dom}) \gets \mtt{pick\_out}(\mtt{doms})$
        \State $\left[ \mtt{subdom\_1}, \dots, \mtt{subdom\_s} \right] \gets \mtt{split}(\mtt{dom})$
        \For{$i = 1 \dots s $}
        \State $\mtt{dom\_ub} \gets \mtt{compute\_UB}(\mtt{net}, \mtt{subdom\_i})$
        \State $\mtt{dom\_lb} \gets \mtt{compute\_LB}(\mtt{net}, \mtt{subdom\_i})$
        \If{$\mtt{dom\_ub} < \mtt{global\_ub}$}
        \State $\mtt{global\_ub} \gets \mtt{dom\_ub}$
        \State $\mtt{prune\_domains}(\mtt{doms}, \mtt{global\_ub})$
        \EndIf
        \If{$\mtt{dom\_lb} < \mtt{global\_ub}$}
        \State $\mtt{domains}.\mtt{append}((\mtt{dom\_lb}, \mtt{subdom\_i}))$
        \EndIf
        \EndFor
        \State $\mtt{global\_lb} \gets \min\{ \mtt{lb}\ |\ (\mtt{lb}, \mtt{dom}) \in \mtt{doms}\}$
        \EndWhile
        \State\Return $\mtt{global\_ub}$
        \EndFunction
      \end{algorithmic}
    \end{algorithm}
  \end{minipage}
\end{wrapfigure}

The description of the verification problem as optimization and the pseudo-code
of Algorithm~\ref{alg:bab} are generic and would apply to verification problems
beyond the specific case of PL-NN. To obtain a practical algorithm, it is
necessary to specify several elements.

\noindent{\bf A search strategy}, defined by the $\mtt{pick\_out}$ function, which
chooses the next domain to branch on. Several heuristics are possible, for
example those based on the results of previous bound computations. For
satisfiable problems or optimization problems, this allows to discover good
upper bounds, enabling early pruning.

\noindent{\bf A branching rule}, defined by the $\mtt{split}$ function, which takes a
domain $\mtt{dom}$ and return a partition in subdomain such that
\mbox{$\bigcup_i \mtt{subdom\_i} = \mtt{dom}$} and that \mbox{$\left(
    \mtt{subdom\_i} \cap \mtt{subdom\_j} \right)= \emptyset, \ \forall i \neq
  j$}. This will define the ``shape'' of the domains, which impacts the hardness
of computing bounds. In addition, choosing the right partition can greatly
impact the quality of the resulting bounds.

\noindent{\bf Bounding methods}, defined by the $\mtt{compute\_\{UB,LB\}}$
functions. These procedures estimate respectively upper bounds and lower bounds
over the minimum output that the network $\mtt{net}$ can reach over a given
input domain. We want the lower bound to be as high as possible, so that this
whole domain can be pruned easily. This is usually done by introducing convex
relaxations of the problem and minimising them. On the other hand, the computed
upper bound should be as small as possible, so as to allow pruning out other
regions of the space or discovering counterexamples. As any feasible point
corresponds to an upper bound on the minimum, heuristic methods are sufficient.

We now demonstrate how some published work in the literature can be understood
as special case of the branch-and-bound framework for verification.

\subsection{Reluplex}
\label{subsec:reluplex}
\citet{Katz2017} present a procedure named Reluplex to verify properties of
Neural Network containing linear functions and ReLU activation unit, functioning
as an SMT solver using the splitting-on-demand framework~\citep{Barrett2006}. The
principle of Reluplex is to always maintain an assignment to all of the
variables, even if some of the constraints are violated.

Starting from an initial assignment, it attempts to fix some violated
constraints at each step. It prioritises fixing linear constraints
(\eqref{eq:ctx-bounds}, \eqref{eq:lin-op}, \eqref{eq:final} and some relaxation
of \eqref{eq:relu-op}) using a simplex algorithm, even if it leads to violated
ReLU constraints. If no solution to this relaxed problem containing only linear
constraints exists, the counterexample search is unsatisfiable. Otherwise,
either all ReLU are respected, which generates a counterexample, or Reluplex
attempts to fix one of the violated ReLU; potentially leading to newly violated
linear constraints. This process is not guaranteed to converge, so to make
progress, non-linearities that get fixed too often are split into two cases. Two
new problems are generated, each corresponding to one of the phases of the ReLU.
In the worst setting, the problem will be split completely over all possible
combinations of activation patterns, at which point the sub-problems will all be
simple LPs.

This algorithm can be mapped to the special case of branch-and-bound for
satisfiability. The {\bf search strategy} is handled by the SMT core and to the
best of our knowledge does not prioritise any domain. The {\bf branching rule} is
implemented by the ReLU-splitting procedure: when neither the upper bound
search, nor the detection of infeasibility are successful, one non-linear
constraint over the $j$-th neuron of the $i$-th layer \mbox{$x_{i[j]} =
  \max\left( \xhat_{i[j]}, 0 \right)$} is split out into two subdomains:
\mbox{$\{x_{i[j]}=0, \xhat_{i[j]}\leq 0\}$} and \mbox{$\{x_{i[j]}=\xhat_{i[j]},
  \xhat_{i[j]}\geq 0\}$}. This defines the type of subdomains produced. The
prioritisation of ReLUs that have been frequently fixed is a heuristic to decide
between possible partitions.

As Reluplex only deal with satisfiability, the analogue of the lower bound
computation is an over-approximation of the satisfiability problem. The {\bf
  bounding method} used is a convex relaxation, obtained by dropping some of the
constraints. The following relaxation is applied to ReLU units for which the sign of the
input is unknown \mbox{($l_{i[j]} \leq 0$ and $u_{i[j]} \geq 0$)}.

\begin{subequations}
  \begin{minipage}{.45\textwidth}
    \noindent\begin{equation}
    \mbf{x_i} = \max\left(\mbf{\xhat_i}, 0\right)\quad \Rightarrow \quad  \mbf{x_i} \geq
    \mbf{\xhat_i} \label{eq:rpx-reluout-lbinp}
    \end{equation}
  \end{minipage}%
  \begin{minipage}{.25\textwidth}
    \noindent\begin{equation}
    \mbf{x_i} \geq 0 \label{eq:rpx-reluout-lb0}
    \end{equation}
  \end{minipage}%
  \begin{minipage}{.25\textwidth}
    \noindent\begin{equation}
    \mbf{x_i} \leq \mbf{u_i}. \label{eq:rpx-reluout-ub}
    \end{equation}
  \end{minipage}
  \label{eq:reluplex-relax}%
\end{subequations}

\noindent If this relaxation is unsatisfiable, this indicates that the
subdomain cannot contain any counterexample and can be pruned out. The search for
an assignment satisfying all the ReLU constraints by iteratively attempting to
correct the violated ReLUs is a heuristic that is equivalent to the search for an upper
bound lower than 0: success implies the end of the procedure but no guarantees
can be given.

\subsection{Planet}
\label{subsec:planet}
\citet{Ehlers2017} also proposed an approach based on SMT. Unlike Reluplex, the
proposed tool, named Planet, operates by explicitly attempting to find an
assignment to the phase of the non-linearities. Reusing the notation of
Section~\ref{subsec:mip}, it assigns a value of 0 or 1 to each $\delta_{i[j]}$
variable, verifying at each step the feasibility of the partial assignment so as
to prune infeasible partial assignment early.

As in Reluplex, the {\bf search strategy} is not explicitly encoded and simply
enumerates all the domains that have not yet been pruned. The {\bf branching rule} is the
same as for Reluplex, as fixing the decision variable $\delta_{i[j]}=0$ is
equivalent to choosing \mbox{$\{x_{i[j]}=0, \xhat_{i[j]}\leq 0\}$} and fixing
$\delta_{i[j]}=1$ is equivalent to \mbox{$\{x_{i[j]}=\xhat_{i[j]},
  \xhat_{i[j]}\geq 0\}$} . Note however that Planet does not include any heuristic to
prioritise which decision variables should be split over.

Planet does not include a mechanism for early termination based on a heuristic
search of a feasible point. For satisfiable problems, only when a full complete
assignment is identified is a solution returned. In order to detect incoherent
assignments, \citet{Ehlers2017} introduces a global linear approximation to a neural
network, which is used as a {\bf bounding method} to over-approximate the set of
values that each hidden unit can take. In addition to the existing
linear constraints (\eqref{eq:ctx-bounds}, \eqref{eq:lin-op} and
\eqref{eq:final}), the non-linear constraints are approximated by sets of linear
constraints representing the non-linearities' convex hull. Specifically, ReLUs with
input of unknown sign are replaced by the set of equations:

\begin{subequations}
  \begin{minipage}{.40\textwidth}
    \noindent\begin{equation}
      \mbf{x_i} = \max\left(\mbf{\xhat_i}, 0\right)\ \Rightarrow \ \mbf{x_i} \geq \mbf{\xhat_i}
      \label{eq:planet-reluout-lbinp}
    \end{equation}
  \end{minipage}
  \begin{minipage}{.2\textwidth}
    \noindent\begin{equation}
    \mbf{x_i} \geq 0 \label{eq:planet-reluout-lb0}
    \end{equation}
  \end{minipage}
  \begin{minipage}{.35\textwidth}
    \noindent\begin{equation}
      x_{i[j]} \leq u_{i[j]} \frac{\xhat_{i[j]} - l_{i[j]}}{u_{i[j]} - l_{i[j]}}\label{eq:planet-reluout-ub}
    \end{equation}
  \end{minipage}
  \label{eq:planet-relu-hull}%
\end{subequations}
where $x_{i[j]}$ corresponds to the value of the $j$-th coordinate of
$\mbf{x_{i}}$. An illustration of the feasible domain is provided in the
supplementary material.



Compared with the relaxation of Reluplex~\eqref{eq:reluplex-relax}, the Planet
relaxation is tighter. Specifically, Eq.~\eqref{eq:rpx-reluout-lbinp} and
\eqref{eq:rpx-reluout-lb0} are identical to Eq.~\eqref{eq:planet-reluout-lbinp}
and \eqref{eq:planet-reluout-lb0} but Eq.~\eqref{eq:planet-reluout-ub} implies
Eq.~\eqref{eq:rpx-reluout-ub}. Indeed, given that $\xhat_{i[j]}$ is smaller than
$u_{i[j]}$, the fraction multiplying $u_{i[j]}$ is necessarily smaller than 1,
implying that this provides a tighter bounds on $x_{i[j]}$.

To use this approximation to compute better bounds than the ones given by simple
interval arithmetic, it is possible to leverage the feed-forward
structure of the neural networks and obtain bounds one layer at a time. Having
included all the constraints up until the $i$-th layer, it is possible to
optimize over the resulting linear program and obtain bounds for all the units
of the $i$-th layer, which in turn will allow us to create the
constraints~\eqref{eq:planet-relu-hull} for the next layer.

In addition to the pruning obtained by the convex relaxation, both Planet and
Reluplex make use of conflict analysis~\citep{Marques1999} to discover
combinations of splits that cannot lead to satisfiable assignments, allowing
them to perform further pruning of the domains.

\subsection{Potential improvements}
As can be seen, previous approaches to neural network verification have relied
on methodologies developed in three communities: optimization, for the creation
of upper and lower bounds; verification, especially SMT; and machine learning,
especially the feed-forward nature of neural networks for the creation of
relaxations. A natural question that arises is ``Can other existing literature
from these domains be exploited to further improve neural network
verification?''  Our unified branch-and-bound formulation makes it easy to answer
this question. To illustrate its power, we now provide a non-exhaustive list of
suggestions to speed-up verification algorithms.

\paragraph{Better bounding ---} While the relaxation proposed by
\citet{Ehlers2017} is tighter than the one used by Reluplex, it can be improved
further still. Specifically, after a splitting operation, on a smaller domain,
we can refine all the $\mbf{l_i}, \mbf{u_i}$ bounds, to obtain a tither
relaxation. We show the importance of this in the experiments section with the
\textbf{BaB-relusplit} method that performs splitting on the activation like
Planet but updates its approximation completely at each step.

One other possible area of improvement lies in the tightness of the bounds used.
Equation~\eqref{eq:planet-relu-hull} is very closely related to the Mixed
Integer Formulation of Equation~\eqref{eq:mip-form}. Indeed, it corresponds to
level 0 of the Sherali-Adams hierarchy of relaxations~\citep{Sherali1994}. The
proof for this statement can be found in the supplementary material. Stronger
relaxations could be obtained by exploring higher levels of the hierarchy. This
would jointly constrain groups of ReLUs, rather than linearising them
independently.

\paragraph{Better branching }\ The decision to split on the activation of the
ReLU non-linearities made by Planet and Reluplex is intuitive as it provides a
clear set of decision variables to fix. However, it ignores another natural
branching strategy, namely, splitting the input domain. Indeed, it could be
argued that since the function encoded by the neural networks are piecewise
linear in their input, this could result in the computation of highly useful
upper and lower bounds. To demonstrate this, we propose the novel
\textbf{BaB-input} algorithm: a branch-and-bound method that branches over the
input features of the network. Based on a domain with input constrained by
Eq.~\eqref{eq:ctx-bounds}, the $\mtt{split}$ function would return two
subdomains where bounds would be identical in all dimension except for the
dimension with the largest length, denoted $i^\star$. The bounds for each
subdomain for dimension $i^\star$ are given by \mbox{$l_{0[i^\star]} \leq
  x_{0[i^\star]} \leq \frac{l_{0[i^\star]} + u_{0[i^\star]}}{2}$} and
\mbox{$\frac{l_{0[i^\star]} + u_{0[i^\star]}}{2} \leq x_{0[i^\star]} \leq
  u_{0[i^\star]}$}. Based on these tighter input bounds, tighter bounds at all
layers can be re-evaluated.

One of the main advantage of branching over the variables is that all subdomains
generated by the BaB algorithm when splitting over the input variables end up
only having simple bound constraints over the value that input variable can
take. In order to exploit this property to the fullest, we use the highly
efficient lower bound computation approach of \citet{Kolter2017}. This approach
was initially proposed in the context of robust optimization. However, our
unified framework opens the door for its use in verification. Specifically,
\citet{Kolter2017} identified an efficient way of computing bounds for the type
of problems we encounter, by generating a feasible solution to the dual of the
LP generated by the Planet relaxation. While this bound is quite loose compared
to the one obtained through actual optimization, they are very fast to evaluate.
We propose a smart branching method \textbf{BaBSB} to replace the longest edge
heuristic of \textbf{BaB-input}. For all possible splits, we compute fast bounds
for each of the resulting subdomain, and execute the split resulting in the
highest lower bound. The intuition is that despite their looseness, the fast
bounds will still be useful in identifying the promising splits.

\section{Experimental setup}
The problem of PL-NN verification has been shown to be
NP-complete~\citep{Katz2017}. Meaningful comparison between approaches therefore
needs to be experimental.

\subsection{Methods}
The simplest baseline we refer to is \textbf{BlackBox}, a direct encoding of
Eq.~\eqref{eq:sat-problem} into the Gurobi solver, which will perform its own
relaxation, without taking advantage of the problem's structure.

For the SMT based methods, \textbf{Reluplex} and \textbf{Planet}, we use the
publicly available versions~\citep{planetGH,reluplexGH}. Both tools are
implemented in C++ and relies on the GLPK library to solve their relaxation. We
wrote some software to convert in both directions between the input format of
both solvers.

We also evaluate the potential of using MIP solvers, based on the formulation of
Eq.~\eqref{eq:mip-form}. Due to the lack of availability of open-sourced methods
at the time of our experiments, we reimplemented the approach in Python, using
the Gurobi MIP solver. We report results for a variant called
\textbf{MIPplanet}, which uses bounds derived from Planet's convex relaxation
rather than simple interval arithmetic. The MIP is not treated as a simple
feasibility problem but is encoded to minimize the output $\xhat_n$ of
Equation~\eqref{eq:final}, with a callback interrupting the optimization as soon
as a negative value is found. Additional discussions on encodings of the MIP
problem can be found in supplementary materials.

\begin{figure}[h!]
  \begin{minipage}[t]{.52\textwidth}
    \centering
    \vskip 0pt
    \begin{subfigure}[t]{.48\textwidth}
      \includegraphics[width=\textwidth]{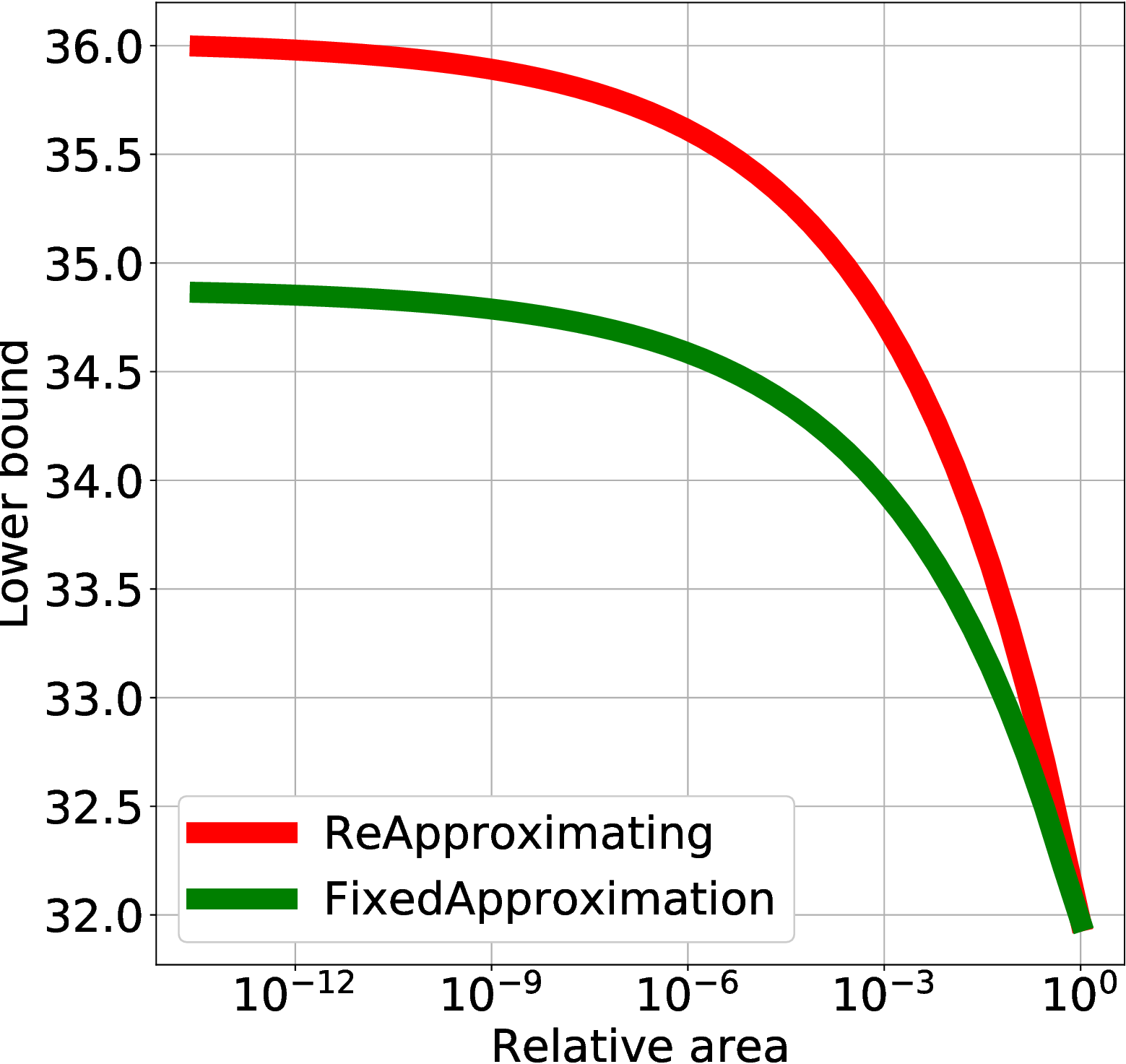}
      \caption{\label{fig:shallow-approx}\small Approximation on a
        \mbox{CollisionDetection} net}
    \end{subfigure}%
    \hfill
    \begin{subfigure}[t]{.48\textwidth}
      \includegraphics[width=\textwidth]{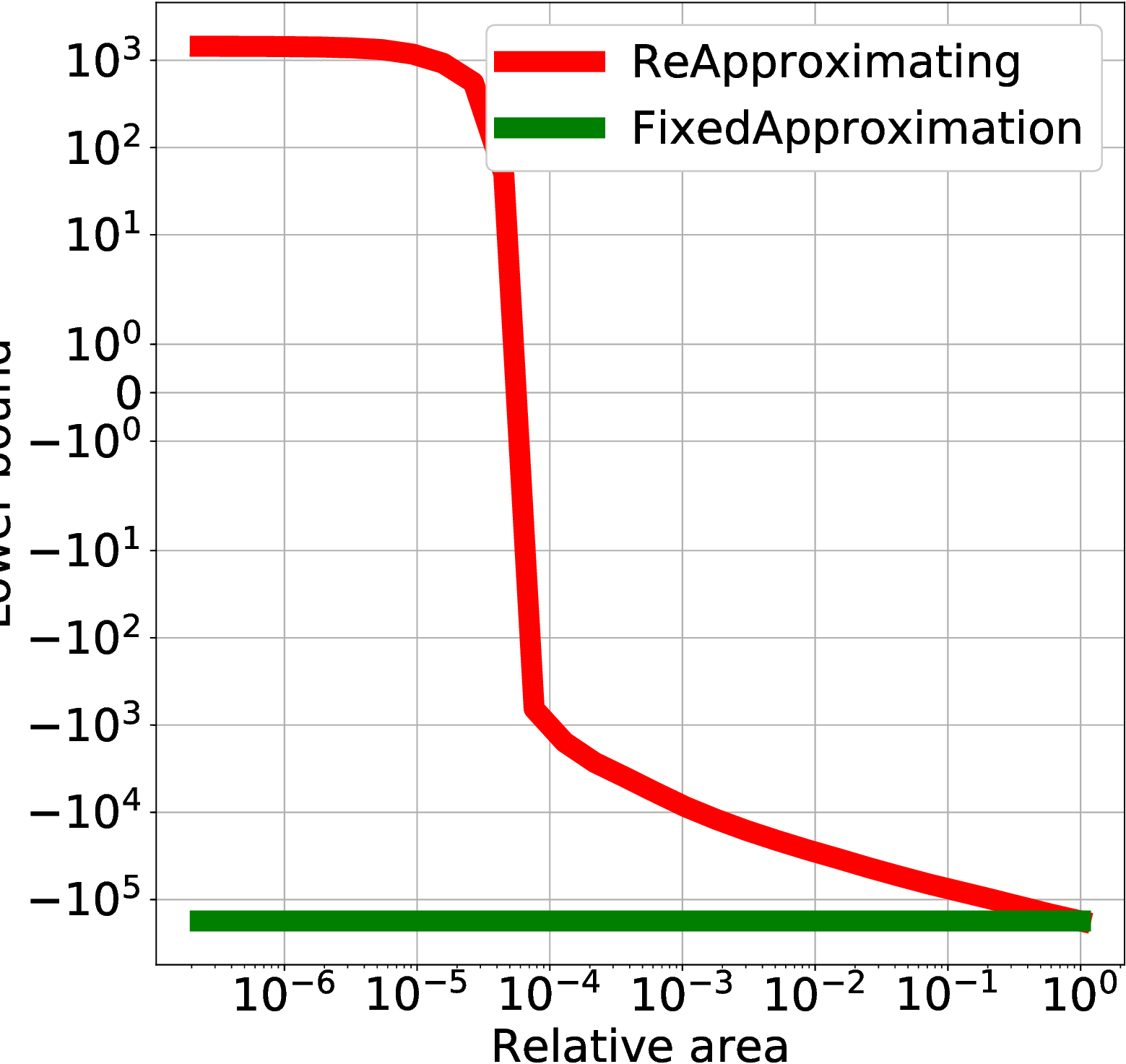}
      \caption{\label{fig:deep-approx}\small Approximation on a deep net from ACAS}
    \end{subfigure}
  \end{minipage}\hfill
  \begin{minipage}[t]{.45\textwidth}
    \caption{\emph{\label{fig:lin-approx}\small Quality of the linear approximation,
        depending on the size of the input domain. We plot the value of the lower
        bound as a function of the area on which it is computed (higher is
        better). The domains are centered around the global minimum and repeatedly
        shrunk. Rebuilding completely the linear approximation at each step allows
        to create tighter lower-bounds thanks to better $\mbf{l_i}$ and
        $\mbf{u_i}$, as opposed to using the same constraints and only changing
        the bounds on input variables. This effect is even more significant on
        deeper networks.}}
  \end{minipage}
  \vspace{-25pt}
\end{figure}

In our benchmark, we include the methods derived from our Branch and Bound
analysis. Our implementation follows faithfully Algorithm~\ref{alg:bab}, is
implemented in Python and uses Gurobi to solve LPs. The $\mtt{pick\_out}$
strategy consists in prioritising the domain that currently has the smallest
lower bound. Upper bounds are generated by randomly sampling points on the
considered domain, and we use the convex approximation of \citet{Ehlers2017} to
obtain lower bounds. As opposed to the approach taken by
\citet{Ehlers2017} of building a single approximation of the network, we rebuild
the approximation and recompute all bounds for each sub-domain. This is
motivated by the observation shown in Figure~\ref{fig:lin-approx} which
demonstrate the significant improvements it brings, especially for deeper
networks. For $\mtt{split}$, \textbf{BaB-input} performs branching by splitting
the input domain in half along its longest edge and \textbf{BaBSB} does it by
splitting the input domain along the dimension improving the global lower bound
the most according to the fast bounds of \citet{Kolter2017}. We also include
results for the \textbf{BaB-relusplit} variant, where the $\mtt{split}$ method
is based on the phase of the non-linearities, similarly to \textbf{Planet}.

\subsection{Evaluation Criteria}
For each of the data sets, we compare the different methods using the same
protocol. We attempt to verify each property with a timeout of two hours, and a
maximum allowed memory usage of 20GB, on a single core of a machine with an
i7-5930K CPU. We measure the time taken by the solvers to either prove or
disprove the property. If the property is false and the search problem is
therefore satisfiable, we expect from the solver to exhibit a counterexample. If
the returned input is not a valid counterexample, we don't count the property as
successfully proven, even if the property is indeed satisfiable. All code and
data necessary to replicate our analysis are released.

\section{Analysis}
We attempt to perform verification on three data sets of properties and report
the comparison results. The dimensions of all the problems can be found in the
supplementary material.

The \textbf{CollisionDetection} data set~\citep{Ehlers2017} attempts to predict
whether two vehicles with parameterized trajectories are going to collide.
500 properties are extracted from problems arising from a binary search to identify
the size of the region around training examples in which the prediction of the
network does not change. The network used is relatively shallow but due to the
process used to generate the properties, some lie extremely close between the
decision boundary between SAT and UNSAT. Results presented in
Figure~\ref{fig:cactus_collision} therefore highlight the accuracy of methods.

The \textbf{Airborne Collision Avoidance System (ACAS)} data set, as released by
\citet{Katz2017} is a neural network based advisory system recommending
horizontal manoeuvres for an aircraft in order to avoid collisions, based on
sensor measurements. Each of the five possible manoeuvres is assigned a score by
the neural network and the action with the minimum score is chosen. The 188
properties to verify are based on some specification describing various
scenarios. Due to the deeper network involved, this data set is useful in
highlighting the scalability of the various algorithms.

\textbf{PCAMNIST} is a novel data set that we introduce to get a better
understanding of what factors influence the performance of various methods. It
is generated in a way to give control over different architecture parameters.
Details of the dataset construction are given in supplementary materials.
We present plots in Figure~\ref{fig:hyp-analysis} showing the
evolution of runtimes depending on each of the architectural parameters of the
networks.

\begin{figure}[h]
  \begin{minipage}[b]{.54\textwidth}
    \vskip 0pt
    \centering
    \begin{subfigure}[t]{.50\textwidth}
      \vskip 0pt
      \includegraphics[width=.90\textwidth]{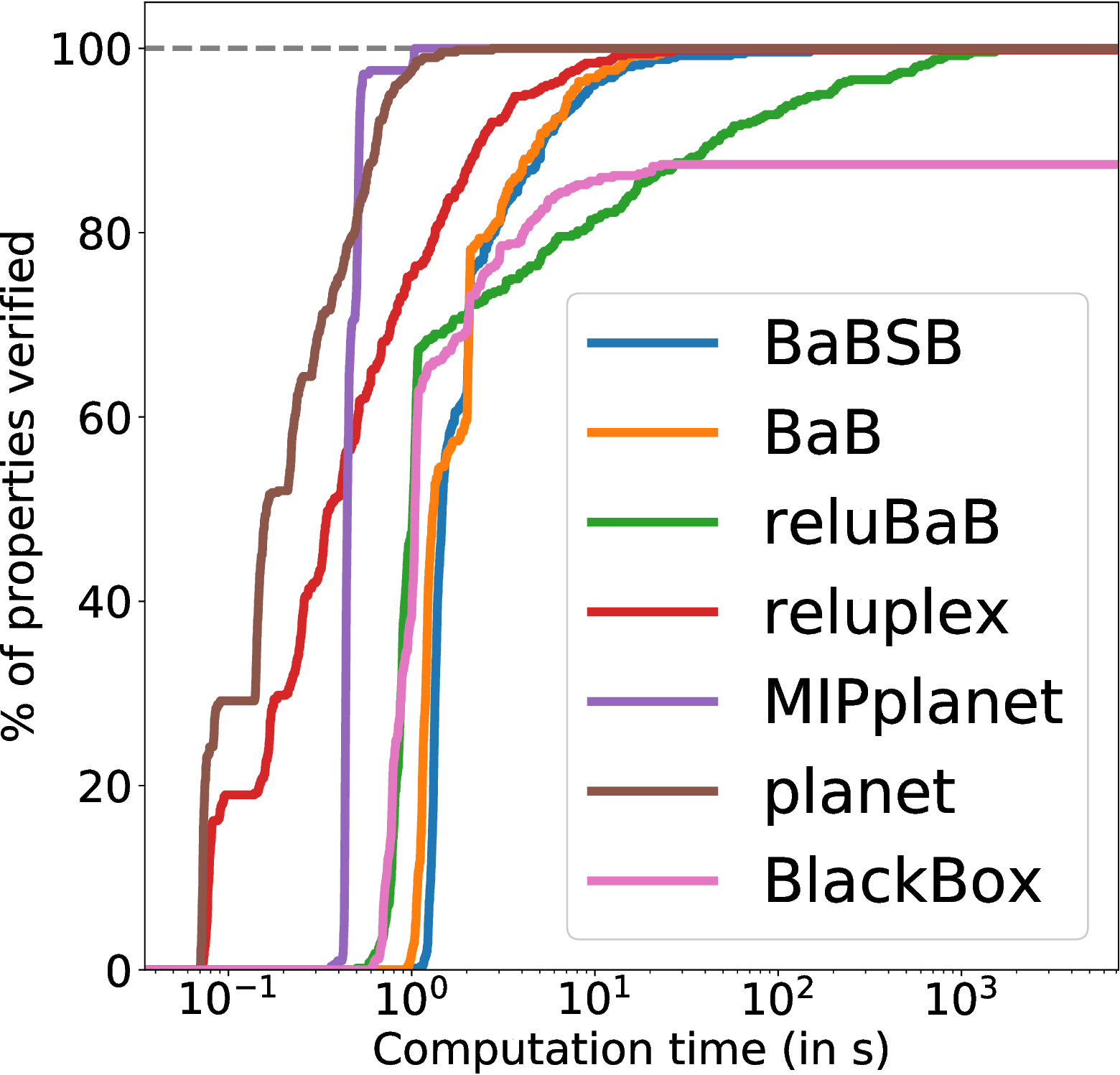}
      \caption{\label{fig:cactus_collision}\emph{CollisionDetection Dataset}}
    \end{subfigure}%
    \begin{subfigure}[t]{.50\textwidth}
      \vskip 0pt
      \includegraphics[width=.90\textwidth]{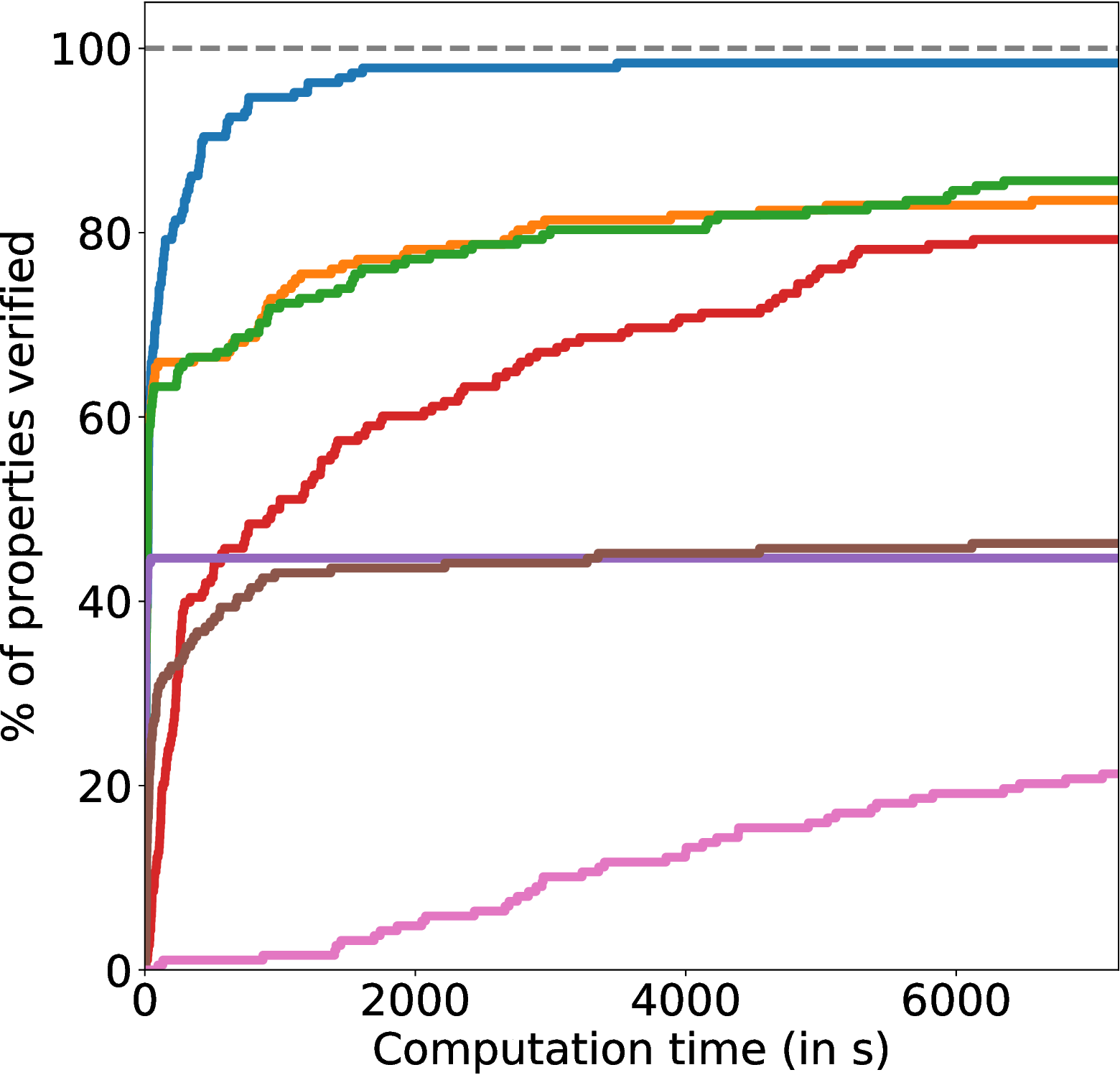}
      \caption{\label{fig:cactus_acas}\emph{ACAS Dataset}}
    \end{subfigure}
  \end{minipage}
  \begin{minipage}[t]{.4\textwidth}
    \vskip 0pt
    \caption{\label{fig:cactus_collision}\emph{ \small Proportion of properties
        verifiable for varying time budgets depending on the methods employed. A
        higher curve means that for the same time budget, more properties will
        be solvable. All methods solve CollisionDetection quite quickly except
        \textbf{reluBaB}, which is much slower and \textbf{BlackBox} who
        produces several incorrect counterexamples.}}
  \end{minipage}
  \vspace{-10pt}
\end{figure}

\begin{figure}[h]
  \begin{minipage}[c]{.25\textwidth}
    \vskip 0pt
    \resizebox{.9\textwidth}{!}{
    \begin{tabular}{lc}
      \toprule
      \textbf{Method} & \begin{tabular}{c}\textbf{Average}\\ \textbf{time}\\ \textbf{per Node}\end{tabular} \\
      \midrule
      BaBSB & 1.81s \\
      BaB & 2.11s \\
      reluBaB & 1.69s \\
      \midrule
      reluplex & 0.30s \\
      \midrule
      MIPplanet & 0.017s \\
      \midrule
      planet & 1.5e-3s \\
      \bottomrule
    \end{tabular}}
    \captionof{table}{\label{tab:nb_node_average}\emph{\small Average time to explore a node for
      each method.}}
  \end{minipage}
  \hfill
  \begin{minipage}[c]{.30\textwidth}
    \includegraphics[width=.95\textwidth]{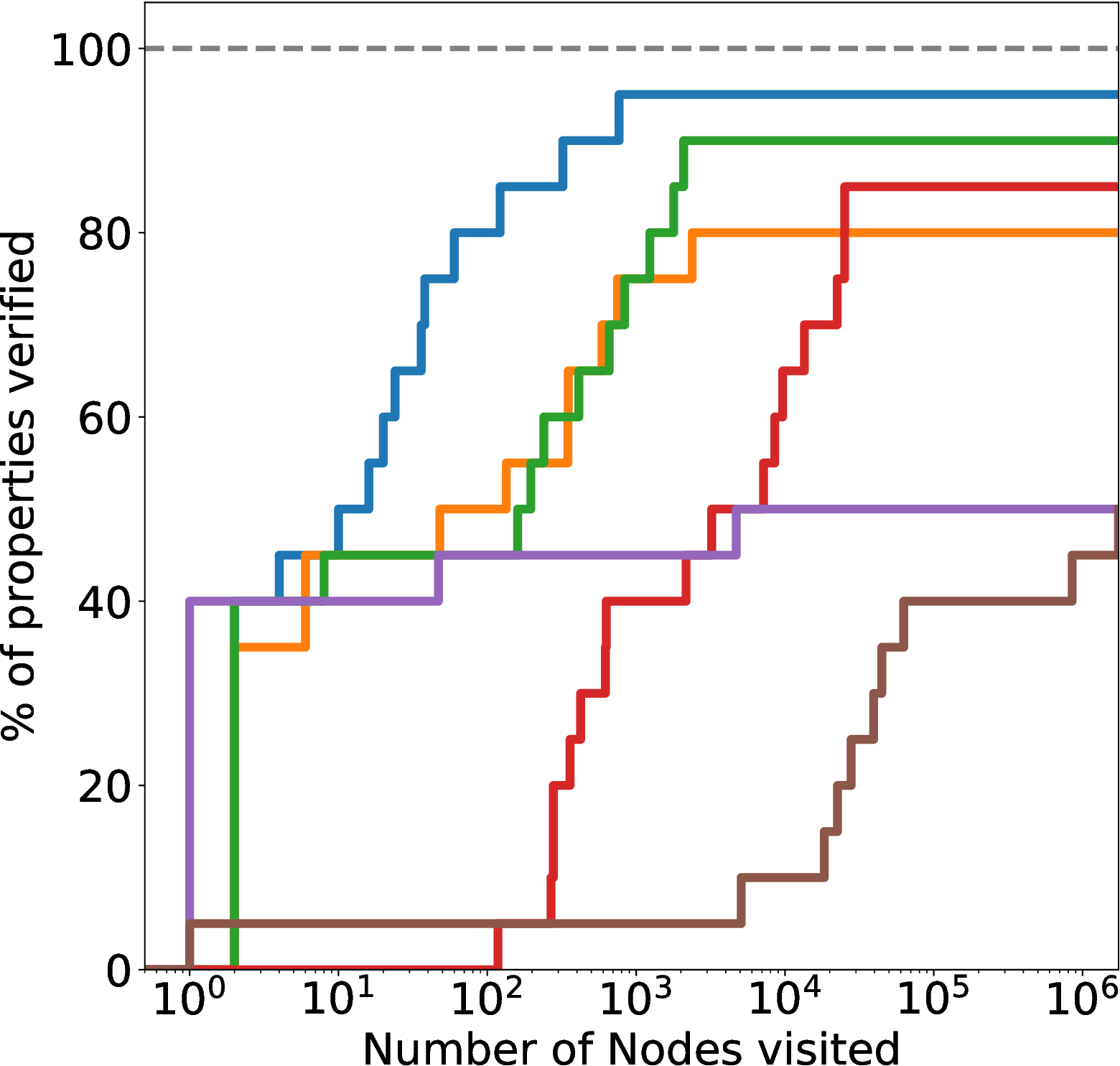}
    \begin{subfigure}{\textwidth}
    \caption{\label{fig:cactus_nbnode}\emph{Properties solved for a given number of
        nodes to explore \ \ \ \ \ \ (log scale).}}
    \end{subfigure}
  \end{minipage}
  \hfill
  \begin{minipage}[c]{.40\textwidth}
    \caption{\label{fig:nb_node_analysis}\emph{\small The trade-off taken by the
        various methods are different. Figure \ref{fig:cactus_nbnode} shows how
        many subdomains needs to be explored before verifying properties while
        Table \ref{tab:nb_node_average} shows the average time cost of exploring
        each subdomain. Our methods have a higher cost per node but they require
        significantly less branching, thanks to better bounding. Note also that
        between \textbf{BaBSB} and \textbf{BaB}, the smart branching reduces by
        an order of magnitude the number of nodes to visit.}}
  \end{minipage}
  \vspace{-20pt}
\end{figure}

\paragraph{Comparative evaluation of verification approaches --- }\
\noindent In Figure~\ref{fig:cactus_collision}, on the shallow networks of
\textbf{CollisionDetection}, most solvers succeed against all properties in
about 10s. In particular, the SMT inspired solvers \textbf{Planet},
\textbf{Reluplex} and the MIP solver are extremely fast. The \textbf{BlackBox}
solver doesn't reach 100\% accuracy due to producing incorrect counterexamples.
Additional analysis can be found about the cause of those errors in the
supplementary materials.

On the deeper networks of \textbf{ACAS}, in Figure~\ref{fig:cactus_acas}, no
errors are observed but most methods timeout on the most challenging testcases.
The best baseline is \textbf{Reluplex}, who reaches 79.26\% success rate at the two hour
timeout, while our best method, \textbf{BaBSB}, already achieves 98.40\% with a
budget of one hour. To reach Reluplex's success rate, the required
runtime is two orders of magnitude smaller.

\paragraph{Impact of each improvement --- }
\noindent To identify which changes allow our method to have good performance, we perform
an ablation study and study the impact of removing some components of our
methods. The only difference between \textbf{BaBSB} and \textbf{BaB} is the
smart branching, which represents a significant part of the performance gap.

Branching over the inputs rather than over the activations does not contribute
much, as shown by the small difference between \textbf{BaB} and
\textbf{reluBaB}. Note however that we are able to use the fast methods
of~\citet{Kolter2017} for the smart branching because branching over the inputs
makes the bounding problem similar to the one solved in robust optimization.
Even if it doesn't improve performance by itself, the new type of split enables
the use of smart branching.

The rest of the performance gap can be attributed to using a better bounds:
\textbf{reluBaB} significantly outperforms \textbf{planet} while using the same
branching strategy and the same convex relaxations. The improvement comes from
the benefits of rebuilding the approximation at each step shown in
Figure~\ref{fig:lin-approx}.

Figure~\ref{fig:nb_node_analysis} presents some additional analysis on a
20-property subset of the ACAS dataset, showing how the methods used impact the
need for branching. Smart branching and the use of better lower bounds reduce
heavily the number of subdomains to explore.

\begin{figure}[h]
  \vspace{-5pt}
  \centering
  \begin{subfigure}{.22\textwidth}
    \includegraphics[width=\textwidth]{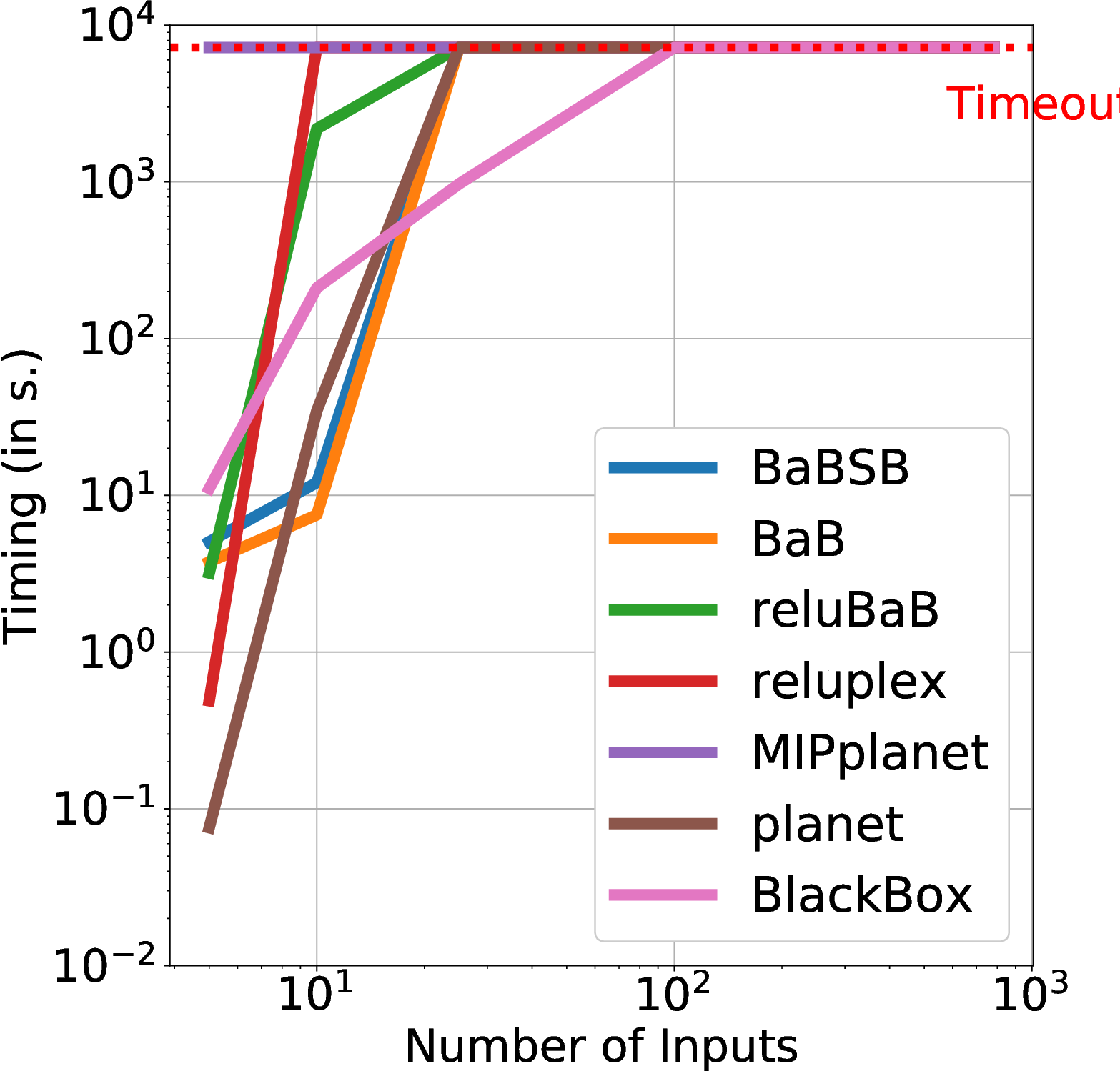}
    \caption{\label{fig:inp-ana}Number of inputs}
    \vspace{-5pt}
  \end{subfigure}%
  \hfill
  \begin{subfigure}{.22\textwidth}
    \includegraphics[width=\textwidth]{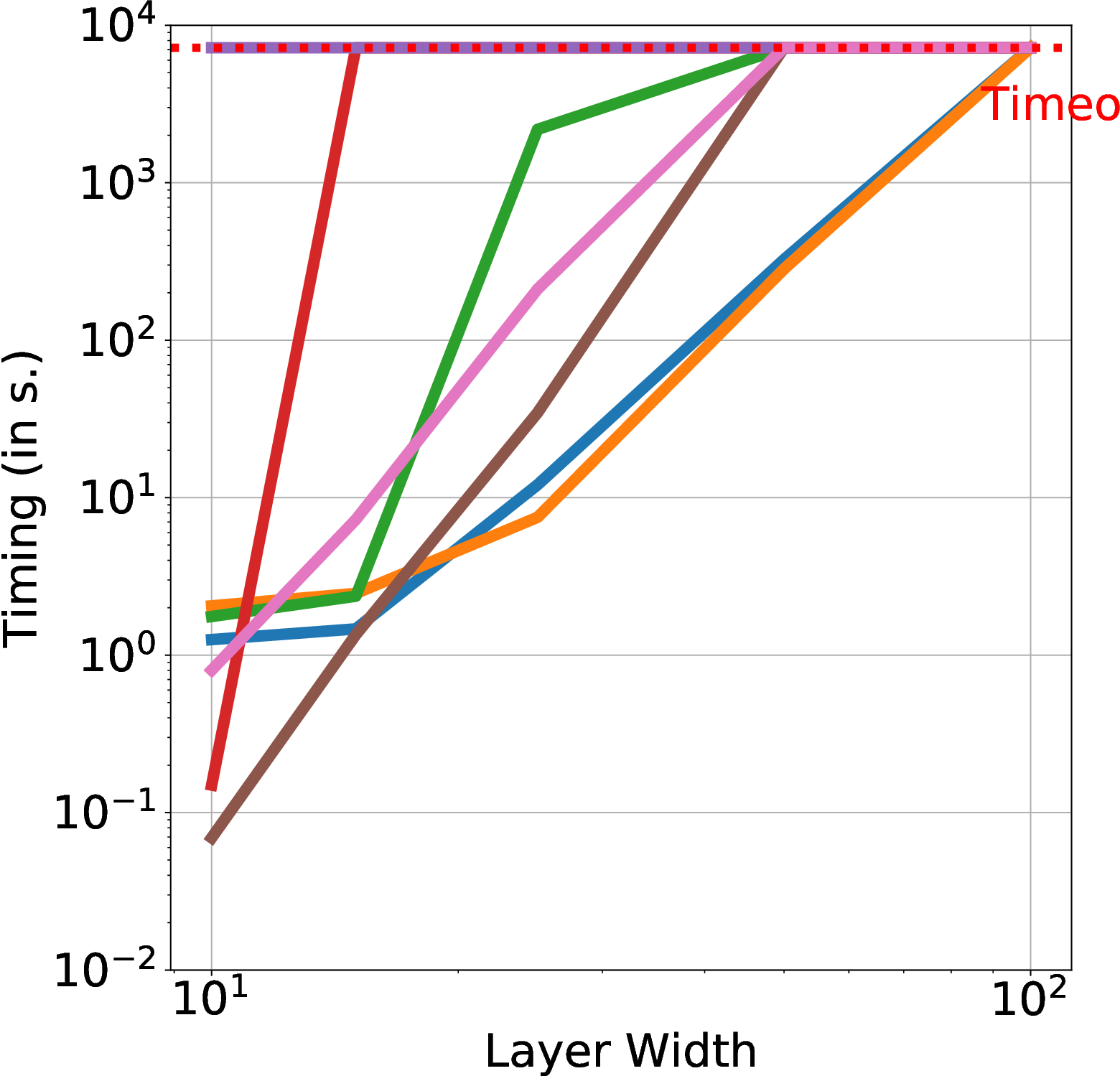}
    \caption{\label{fig:width-ana}Layer width}
    \vspace{-5pt}
  \end{subfigure}%
  \hfill
  \begin{subfigure}{.22\textwidth}
    \includegraphics[width=\textwidth]{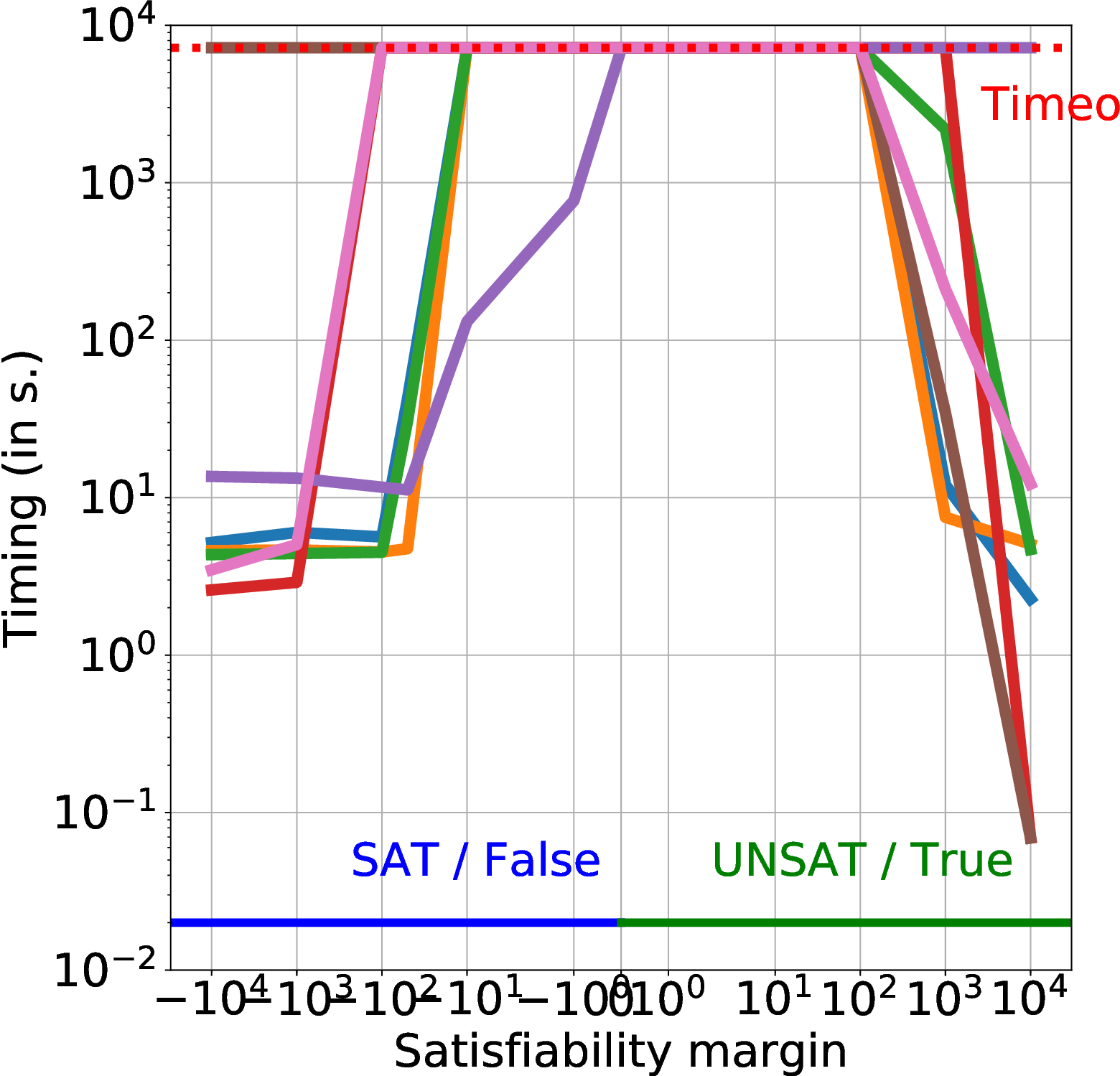}
    \caption{\label{fig:margin-ana}Margin}
    \vspace{-5pt}
  \end{subfigure}%
  \hfill
  \begin{subfigure}{.22\textwidth}
    \includegraphics[width=\textwidth]{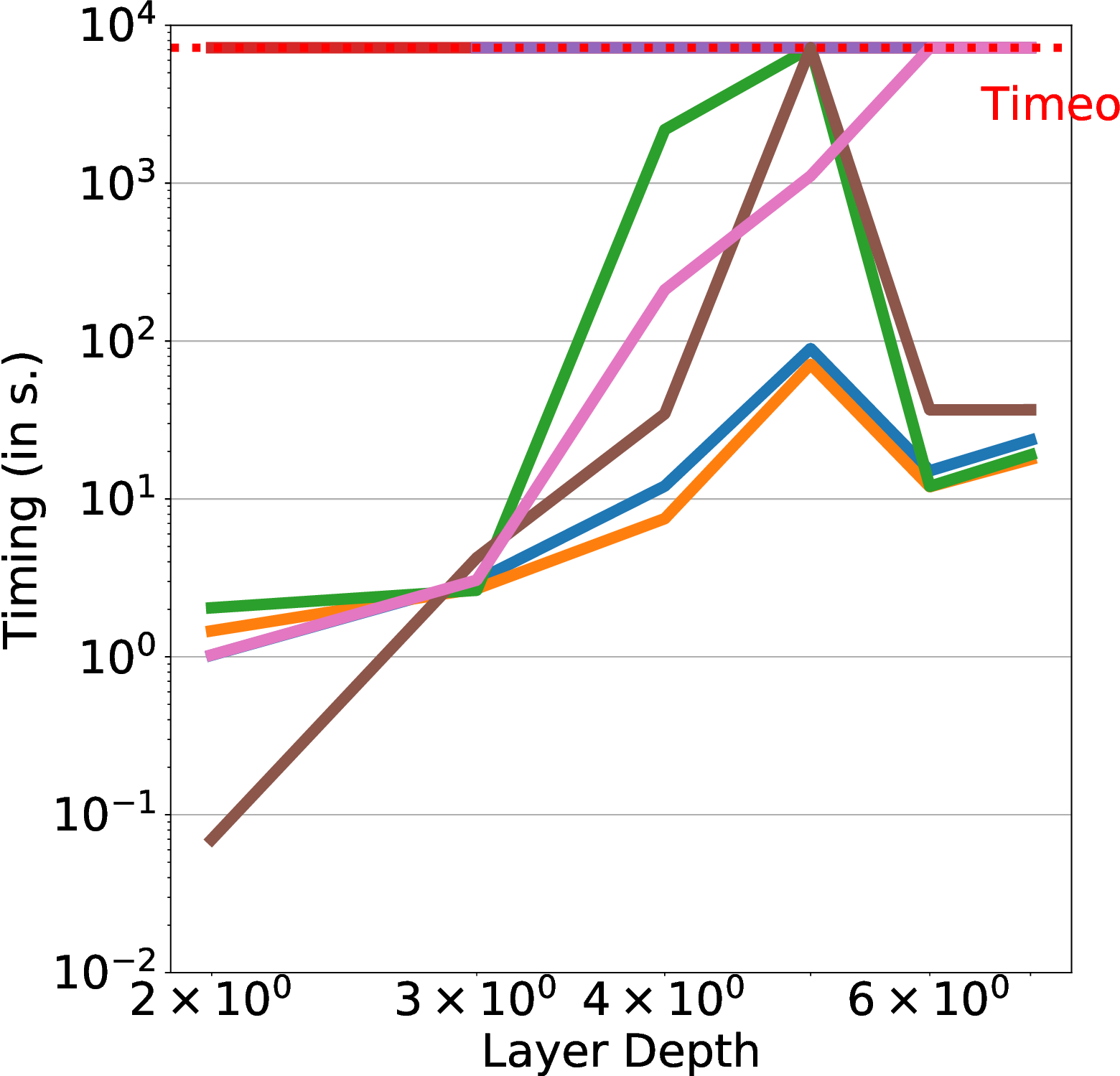}
    \caption{\label{fig:depth-ana}Network depth}
    \vspace{-5pt}
  \end{subfigure}%
  \caption{\label{fig:hyp-analysis} \emph{ \small Impact of the various parameters over
    the runtimes of the different solvers. The base network has 10 inputs and 4
    layers of 25 hidden units, and the property to prove is True with a margin
    of 1000. Each of the plot correspond to a variation of one of this
    parameters.}}
  \vspace{-10pt}
\end{figure}

In the graphs of Figure~\ref{fig:hyp-analysis}, the trend for all the methods
are similar, which seems to indicate that hard properties are intrinsically hard
and not just hard for a specific solver. Figure~\ref{fig:inp-ana} shows an
expected trend: the largest the number of inputs, the harder the problem is.
Similarly, Figure~\ref{fig:width-ana} shows unsurprisingly that wider networks
require more time to solve, which can be explained by the fact that they have
more non-linearities. The impact of the margin, as shown in
Figure~\ref{fig:margin-ana} is also clear. Properties that are True or False
with large satisfiability margin are easy to prove, while properties that have
small satisfiability margins are significantly harder.

\section{Conclusion}
The improvement of formal verification of Neural Networks represents an
important challenge to be tackled before learned models can be used in safety
critical applications. By providing both a unified framework to reason about
methods and a set of empirical benchmarks to measure performance with, we hope
to contribute to progress in this direction. Our analysis of published
algorithms through the lens of Branch and Bound optimization has already
resulted in significant improvements in runtime on our benchmarks. Its continued
analysis should reveal even more efficient algorithms in the future.

\clearpage
\bibliography{shortstrings,oval,paper_specific}
\bibliographystyle{icml2018}

\clearpage
\appendix
\section{Canonical Form of the verification problem}

If the property is a simple inequality \mbox{$P(\mbf{\xhat_n}) \triangleq
  \mbf{c}^T \mbf{\xhat_n} \geq b$}, it is sufficient to add to the network a
final fully connected layer with one output, with weight of $\mbf{c}$ and a bias
of $-b$. If the global minimum of this network is positive, it indicates that
for all $\mbf{\xhat_n}$ the original network can output, we have
\mbox{$\mbf{c}^T \mbf{\xhat_n} -b \geq 0 \implies \mbf{c}^T \mbf{\xhat_n} \geq
  b$}, and as a consequence the property is True. On the other hand, if the
global minimum is negative, then the minimizer provides a counter-example.

Clauses specified using \texttt{OR} (denoted by $\bigvee$) can be encoded by
using a MaxPooling unit. If the property is \mbox{$P(\mbf{\xhat_n}) \triangleq
  \bigvee_i\left[ \mbf{c}_i^T \mbf{\xhat_n} \geq b_i \right]$}, this is
equivalent to $\max_i\left( \mbf{c}_i^T \mbf{\xhat_n} - b_i \right) \geq 0$.

Clauses specified using \texttt{AND} (denoted by $\bigwedge$) can be encoded
similarly: the property $P(\xhat_n) = \bigwedge_i \left[
  \mbf{c}_i^T\mbf{\xhat_n} \geq b_i \right]$ is equivalent to
\mbox{\small$\min_i \left( \mbf{c}_i^T \mbf{\xhat_n} - b_i \right) \geq 0 \iff
  -\left( \max_i\left(-\mbf{c}_i^T \mbf{\xhat_n} + b_i \right) \right) \geq 0$}

\section{Toy Problem example}
We have specified the problem of formal verification of neural networks as
follows: given a network that implements a function $\mbf{\xhat_n} =
f(\mbf{x_0})$, a bounded input domain $\mathcal{C}$ and a property $P$, we want
to prove that
\begin{equation}
  \mbf{x_0} \in \mathcal{C},\quad \mbf{\xhat_n} = f(\mbf{x_0}) \implies P(\mbf{\xhat_n}).
  \label{eq:prob-form}
\end{equation}

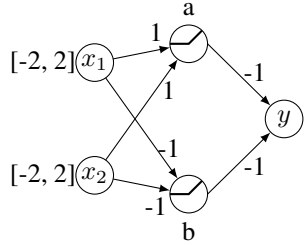
\begin{figure}[h]
  \centering
  \begin{tikzpicture}

  \tikzset{dummy/.style= {inner sep=0, outer sep=0}}
  \tikzset{neuron/.style={draw, circle,inner sep=0, outer sep=0, minimum size=0.5cm}};

  \node[neuron](x) {$x_1$};
  \node[dummy, left = 0 of x](x_range) {[-2, 2]};
  \node[neuron, below = of x](y) {$x_2$};
  \node[dummy, left = 0 of y](y_range) {[-2, 2]};

  \node[neuron, right = of x.north](a) {};
  \draw[black, thick](a.west) to (a.center) to (a.north east);
  \node[above = 0 of a](alabel) {a};
  \node[neuron, right = of y.south](b) {};
  \draw[black, thick](b.west) to (b.center) to (b.north east);
  \node[below = 0 of b](blabel) {b};

  \draw[-latex] (x) to node[near end, above](Wxa){1} (a);
  \draw[-latex] (y) to node[very near end, below](Wya){1} (a);
  \draw[-latex] (x) to node[very near end, above](Wxb){-1} (b);
  \draw[-latex] (y) to node[near end, below](Wyb){-1} (b);

  \node[dummy](inp-mid) at ($(a)!0.5!(b)$){};
  \node[neuron, right = of inp-mid](out) {$y$};

  \draw[-latex] (a.east) to node[near end, above](Wao){-1} (out);
  \draw[-latex] (b.east) to node[near end, below](Wbo){-1} (out);

  \node[draw, thick, below = of b](property) {\textbf{Prove that $y > -5$}};
\end{tikzpicture}
  \caption{\label{fig:net-example} Example Neural Network. We attempt to prove
    the property that the network output is always greater than -5}
\end{figure}

A toy-example of the Neural Network verification problem is given in
Figure~\ref{fig:net-example}. On the domain $\mathcal{C} = [-2; 2] \times [-2;
2]$, we want to prove that the output $y$ of the one hidden-layer network
always satisfy the property $P(y) \triangleq \left[ y > -5 \right]$. We will use
this as a running example to explain the methods used for comparison in our
experiments.

\subsection{Problem formulation}
For the network of Figure~\ref{fig:net-example}, the variables would be $\{x_1,
x_2, a_\text{in}, a_\text{out}, b_\text{in}, b_\text{out}, y\}$ and the set of
constraints would be:

\begin{subequations}
  \begin{align}
    & -2 \leq x_1 \leq 2 \qquad& -2 \leq x_2 \leq 2  \label{eq:ex-bounds}\\
    & \hat{a} = x_1 + x_2  \qquad& \hat{b} = -x_1 - x_2  \nonumber\\
    &y = -a - b  & \label{eq:ex-lin}\\
    & a = \max(\hat{a}, 0)  \qquad& b = \max(\hat{b}, 0) \label{eq:ex-nonlin} \\
    &  y \leq -5  & \label{eq:ex-final}
  \end{align}
\end{subequations}

Here, $\hat{a}$ is the input value to hidden unit, while $a$ is the value after
the ReLU. Any point satisfying all the above constraints would be a
counterexample to the property, as it would imply that it is possible to drive
the output to -5 or less.

\subsection{MIP formulation}
In our example, the non-linearities of equation~\eqref{eq:ex-nonlin} would be
replaced by
\begin{equation}
  \begin{split}
    a \geq 0 \qquad\qquad& a \geq \hat{a} \\
    a \leq \hat{a} - l_a (1-\delta_a) \qquad\qquad& a \leq u_a \delta_a \\
    \delta_a \in \{0, 1\}
  \end{split}
  \label{eq:mip-relu}
\end{equation}
where $l_a$ is a lower bound of the value that $\hat{a}$ can take (such as -4)
and $u_a$ is an upper bound (such as 4). The binary variable $\delta_a$
indicates which phase the ReLU is in: if $\delta_a=0$, the ReLU is blocked and
$a_\text{out}=0$, else the ReLU is passing and $a_\text{out} = a_\text{in}$. The
problem remains difficult due to the integrality constraint on $\delta_a$.

\subsection{Running Reluplex}
Table~\ref{tab:reluplex-ex} shows the initial steps of a run of the Reluplex
algorithm on the example of Figure~\ref{fig:net-example}. Starting from an
initial assignment, it attempts to fix some violated constraints at each step.
It prioritises fixing linear constraints (\eqref{eq:ex-bounds},
\eqref{eq:ex-lin} and \eqref{eq:ex-final} in our illustrative example) using a simplex algorithm, even
if it leads to violated ReLU constraints~\eqref{eq:ex-nonlin}. This can be seen
in step 1 and 3 of the process.

\begin{figure}
  \centering
  {\small
    \begin{tabular}{?c?c|c|c|c|c|c|c?}
      \specialrule{1.5pt}{1pt}{1pt}
      \textbf{Step}& $\mbf{x_1}$ & $\mbf{x_2}$ & $\mbf{\hat{a}}$ & $\mbf{a}$ & $\mbf{\hat{b}}$ & $\mbf{b}$ & $\mbf{y}$ \\
      \specialrule{1.5pt}{1pt}{1pt}
      \multirow{3}{1em}{1}&0 & 0 & 0 & 0 & 0 & 0 & \cellcolor{red}0 \\[0.1ex]
      & \multicolumn{7}{c?}{Fix linear constraints}\\
      & 0 & 0 & \cellcolor{orange}0 & \cellcolor{orange}1 & \cellcolor{orange}0 & \cellcolor{orange}4& -5 \\
      \specialrule{1.5pt}{1pt}{1pt}
      \multirow{3}{1em}{2}& 0 & 0 & \cellcolor{orange}0 & \cellcolor{orange}1 & \cellcolor{orange}0 & \cellcolor{orange}4 & -5 \\
      & \multicolumn{7}{c?}{Fix a ReLU}\\
      & \cellcolor{red}0 & \cellcolor{red}0 & \cellcolor{orange}0 & \cellcolor{orange}1 & \cellcolor{red}4 & 4 & -5 \\
      \specialrule{1.5pt}{1pt}{1pt}
      \multirow{3}{1em}{3} & \cellcolor{red}0 & \cellcolor{red}0 & \cellcolor{orange}0 & \cellcolor{orange}1 & \cellcolor{red}4 & 4 & -5 \\
      & \multicolumn{7}{c|}{Fix linear constraints}\\
      & -2 & -2 & \cellcolor{orange}-4 & \cellcolor{orange}1 & 4 & 4 & -5 \\
      \specialrule{1.5pt}{1pt}{1pt}
      & \multicolumn{7}{|c?}{$\dots$}\\
   \end{tabular}
 }
 \caption{\label{tab:reluplex-ex}\emph{Evolution of the Reluplex algorithm. Red
   cells corresponds to value violating Linear constraints, and orange cells
   corresponds to value violating ReLU constraints. Resolution of violation of linear
   constraints are prioritised.}}
\end{figure}

If no solution to the problem containing only linear constraints exists, this
shows that the counterexample search is unsatisfiable. Otherwise, all linear
constraints are fixed and Reluplex attempts to fix one violated ReLU at a time,
such as in step 2 of Table~\ref{tab:reluplex-ex} (fixing the ReLU $b$),
potentially leading to newly violated linear constraints. In the case where no
violated ReLU exists, this means that a satisfiable assignment has been found
and that the search can be interrupted.

This process is not guaranteed to converge, so to guarantee progress,
non-linearities getting fixed too often are split into two cases. This generates
two new sub-problems, each involving an additional linear constraint instead of
the linear one. The first one solves the problem where \mbox{$\hat{a}\leq 0
  \text{ and } a=0$}, the second one where \mbox{$\hat{a} \geq 0 \text{ and } a
  = \hat{a}$}. In the worst setting, the problem will be split completely over
all possible combinations of activation patterns, at which point the
sub-problems are simple LPs.

\section{Planet approximation}
The feasible set of the Mixed Integer Programming formulation is given by the
following set of equations. We assume that all $\mbf{l_i}$ are negative and
$\mbf{u_i}$ are positive. In case this isn't true, it is possible to just update
the bounds such that they are.

\begin{subequations}
  \begin{align}
    &\mbf{l_0} \leq \mbf{x_0} \leq \mbf{u_0}& \label{eq:ctx-bounds}\\
    &\mbf{\xhat_{i+1}} = W_{i+i} \mbf{x_i} + \mbf{b_{i+i}} &\quad \forall i \in \{0, \ n-1\}\label{eq:lin-op}\\
    &\mbf{x_i} \geq 0 &\quad \forall i \in \{1, \ n-1\}\\
    &\mbf{x_i} \geq \mbf{\xhat_i}&\quad \forall i \in \{1, \ n-1\}\\
    &\mbf{x_i} \leq \mbf{\xhat_i} - \mbf{l_i}\cdot(1 - \bm{\delta_i})&\quad \forall i \in \{1, \ n-1\}\\
    &\mbf{x_i} \leq \mbf{u_i} \cdot \bm{\delta_i}&\quad \forall i \in \{1, \ n-1\}\\
    &\bm{\delta_i} \in \{0, 1\}^{h_i}&\quad \forall i \in \{1, \ n-1\}\\
    &\xhat_{n} \leq 0 \label{eq:final}
  \end{align}
  \label{eq:mipsat-problem}
\end{subequations}

The level 0 of the Sherali-Adams hierarchy of relaxation \citet{Sherali1994}
doesn't include any additional constraints. Indeed, polynomials of degree 0 are
simply constants and their multiplication with existing constraints followed by
linearization therefore doesn't add any new constraints. As a result, the
feasible domain given by the level 0 of the relaxation corresponds simply to the
removal of the integrality constraints:
\begin{subequations}
  \begin{align}
    &\mbf{l_0} \leq \mbf{x_0} \leq \mbf{u_0}& \label{eq:ctx-bounds}\\
    &\mbf{\xhat_{i+1}} = W_{i+i} \mbf{x_i} + \mbf{b_{i+i}} &\quad \forall i \in \{0, \ n-1\}\label{eq:lin-op}\\
    &\mbf{x_i} \geq 0 &\quad \forall i \in \{1, \ n-1\}\\
    &\mbf{x_i} \geq \mbf{\xhat_i}&\quad \forall i \in \{1, \ n-1\}\\
    &\mbf{x_i} \leq \mbf{\xhat_i} - \mbf{l_i}\cdot(1 - \bm{d_i})&\quad \forall i \in \{1, \ n-1\}\label{eq:xub1}\\
    &\mbf{x_i} \leq \mbf{u_i} \cdot \bm{d_i}&\quad \forall i \in \{1, \ n-1\}\label{eq:xub2}\\
    & \underline{0 \leq \bm{d_i} \leq 1 }&\quad \forall i \in \{1, \ n-1\}\\
    &\xhat_{n} \leq 0 \label{eq:final}
  \end{align}
  \label{eq:mipsat-problem}
\end{subequations}

Combining the equations \eqref{eq:xub1} and \eqref{eq:xub2}, looking at a single
unit $j$ in layer $i$, we obtain:

\begin{equation}
  x_{i[j]} \leq \min\left( \xhat_{i[j]} - l_i (1 - d_{i[j]}), u_{i[j]} d_{i[j]} \right)
  \label{eq:x-minub}
\end{equation}
The function mapping $d_{i[j]}$ to an upperbound of $x_{i[j]}$ is a minimum of
linear functions, which means that it is a concave function. As one of them is
increasing and the other is decreasing, the maximum will be reached when they
are both equals.

\begin{equation}
  \begin{split}
    \xhat_{i[j]} - l_{i[j]} (1 - d^{\star}_{i[j]}) &= u_{i[j]}d^{\star}_{i[j]}\\
    \Leftrightarrow \qquad d^{\star}_{i[j]} &= \frac{\xhat_{i[j]} - l_{i[j]}}{u_{i[j]}-l_{i[j]}}
  \end{split}
\end{equation}

Plugging this equation for $d^{\star}$ into Equation\eqref{eq:x-minub} gives
that:
\begin{equation}
    x_{i[j]} \leq u_{i[j]} \frac{\xhat_{i[j]} - l_{i[j]}}{u_{i[j]} - l_{i[j]}}
\end{equation}
which corresponds to the upper bound of $x_{i[j]}$ introduced for Planet~\citep{Ehlers2017}.

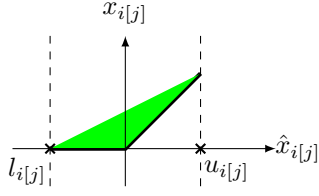
\begin{figure}
  \centering
  \begin{tikzpicture}
  \tikzset{dummy/.style= {inner sep=0, outer sep=0}}
  \tikzset{cross/.style={cross out, draw,
      minimum size=3*(#1-\pgflinewidth),
      inner sep=0pt, outer sep=0pt,
      thick}}

  \draw[-, ultra thick](-1, 0) to (0, 0) to (1, 1);

  \draw[dashed](-1, -0.5) to (-1, 1.5);
  \draw[dashed](1, -0.5) to (1, 1.5);

  \draw[fill=green](-1, 0) -- (0,0) -- (1, 1);

  \node[cross=2pt] at (-1, 0) {};
  \node[dummy](lb-lab) at (-1.3, -0.3) {$l_{i[j]}$};
  \node[cross=2pt] at (1, 0) {};
  \node[dummy](ub-lab) at (1.35, -0.3) {$u_{i[j]}$};

  \draw[-latex](-1.5,0) to (2, 0);
  \node[dummy](x-label) at (2.3, 0) {$\xhat_{i[j]}$};
  \draw[-latex](0,-0.5) to (0, 1.5);
  \node[dummy](x-label) at (0, 1.8) {$x_{i[j]}$};

\end{tikzpicture}
  \caption{\label{fig:relu-hull} \emph{Feasible domain corresponding to the
    Planet relaxation for a single ReLU.}}
\end{figure}

\section{MaxPooling}
For space reason, we only described the case of ReLU activation function in the
main paper. We now present how to handle MaxPooling activation, either by
converting them to the already handled case of ReLU activations or by
introducing an explicit encoding of them when appropriate.

\subsection{Mixed Integer Programming}
Similarly to the encoding of ReLU constraints using binary variables and bounds
on the inputs, it is possible to similarly encode MaxPooling constraints.
The constraint
\begin{equation}
  y = \max\left( x_1, \dots, x_k \right)
\end{equation}
can be replaced by
\begin{subequations}
  \begin{align}
    &y \geq x_i \qquad &&\forall i \in \{1\dots k\}\\
    &y \leq x_i + (u_{x_{1:k}} - l_{x_i})(1 - \delta_i) \quad &&\forall i \in \{1\dots k\}\\
    &\sum_{i \in \{1\dots k\}}  \delta_i = 1 \\
    &\delta_i \in \{0, 1\} \quad &&\forall i \in \{1\dots k\}
  \end{align}
\end{subequations}
where $u_{x_{1:k}}$ is an upper-bound on all $x_{i}$ for $i \in \{1 \dots k\}$ and $l_{x_i}$ is a lower bound on $x_{i}$.

\subsection{Reluplex}
In the version introduced by \citep{Katz2017}, there is no support for
MaxPooling units. As the canonical representation we evaluate needs them, we
provide a way of encoding a MaxPooling unit as a combination of Linear function
and ReLUs.

To do so, we decompose the element-wise maximum into a series of pairwise
maximum
\begin{equation}
  \begin{split}
    \max\left( x_j, x_2, x_3, x_4 \right) = \max(\ &\max\left( x_1, x_2 \right),\\
      & \max\left( x_3, x_4 \right) )
  \end{split}
\end{equation}
and decompose the pairwise maximums as sum of ReLUs:
\begin{equation}
  \max\left( x_1, x_2 \right) = \max\left( x_1 - x_2, \ 0 \right) + \max\left( x_2 - l_{x_2}, 0 \right) + l_{x_2},
\end{equation}
where $l_{x_2}$ is a pre-computed lower bound of the value that $x_2$
can take.

As a result, we have seen that an elementwise maximum such as a MaxPooling unit
can be decomposed as a series of pairwise maximum, which can themselves be
decomposed into a sum of ReLUs units. The only requirement is to be able to
compute a lower bound on the input to the ReLU, for which the methods discussed
in the paper can help.

\subsection{Planet}
As opposed to Reluplex, Planet~\citet{Ehlers2017} directly supports MaxPooling units.
The SMT solver driving the search can split either on ReLUs, by considering
separately the case of the ReLU being passing or blocking. It also has the
possibility on splitting on MaxPooling units, by treating separately each
possible choice of units being the largest one.

For the computation of lower bounds, the constraint
\begin{equation}
  y = \max\left( x_1, x_2, x_3, x_4 \right)
\end{equation}
is replaced by the set of constraints:
\begin{subequations}
  \begin{align}
    y &\geq \text{x}_i \qquad \forall i \in \{1 \dots 4\}\\
    y &\leq \sum_i \left( x_i - l_{x_i} \right) + \max_i l_{x_i},
  \label{eq:maxpool-cvxhull}
  \end{align}
\end{subequations}
where $x_i$ are the inputs to the MaxPooling unit and $l_{x_i}$ their lower
bounds.

\section{Mixed Integers Variants}

\subsection{Encoding}
Several variants of encoding are available to use Mixed Integer Programming as a
solver for Neural Network Verification. As a reminder, in the main paper we used
the formulation of \citet{Tjeng2017}:
\begin{subequations}
  \begin{flalign}
    x_i = \max\left(\mbf{\xhat_i}, 0\right) \quad \Rightarrow \quad \bm{\delta_i} \in
    \{0,1\}^{h_i}, &\quad\mbf{x_i} \geq 0, \ \qquad  \mbf{x_i} \leq \mbf{u_i} \cdot \bm{\delta_i}\label{eq:mip_0}\\
    & \quad \mbf{x_i}\geq \mbf{\xhat_i},   \qquad \mbf{x_i} \leq \mbf{\xhat_i} - \mbf{l_i}\cdot(1 - \bm{\delta_i})\label{eq:mip_inp}
  \end{flalign}
  \label{eq:mip-form}%
\end{subequations}

An alternative formulation is the one of \citet{Lomuscio2017} and \citet{Cheng2017}:
\begin{subequations}
  \begin{flalign}
    x_i = \max\left(\mbf{\xhat_i}, 0\right) \quad \Rightarrow \quad \bm{\delta_i} \in
    \{0,1\}^{h_i}, &\quad\mbf{x_i} \geq 0, \ \qquad  \mbf{x_i} \leq \mbf{M_i} \cdot \bm{\delta_i}\label{eq:mip_0}\\
    & \quad \mbf{x_i}\geq \mbf{\xhat_i},   \qquad \mbf{x_i} \leq \mbf{\xhat_i} - \mbf{M_i}\cdot(1 - \bm{\delta_i})\label{eq:mip_inp}
  \end{flalign}
  \label{eq:altmip-form}%
\end{subequations}
where $\mbf{M_i} = \max\left( -\mbf{l_i}, \mbf{u_i} \right)$. This is
fundamentally the same encoding but with a sligthly worse bounds that is used,
as one of the side of the bounds isn't as tight as it could be.

\subsection{Obtaining bounds}
The formulation described in Equations~\eqref{eq:mip-form} and
~\eqref{eq:altmip-form} are dependant on obtaining lower and upper bounds for
the value of the activation of the network.

\paragraph{Interval Analysis}
One way to obtain them, mentionned in the paper, is the use of interval
arithmetic~\citep{Hickey2001}. If we have bounds $\mbf{l_{i}}, \mbf{u_{i}}$ for a vector
$\mbf{x_{i}}$, we can derive the bounds $\mbf{\hat{l}_{i+1}}, \mbf{\hat{u}_{i+1}}$ for a vector
\mbox{$\mbf{\xhat_{i+1}} = W_{i+1} \mbf{x_i} + b_{i+1}$}

\begin{subequations}
  \begin{align}
    \hat{l}_{i+1[j]} &= \sum_k \left( W_{i+1[j, k]}^+ l^+_{i[k]} + W_{i+1[j,k]}^- u^+_{i[k]} \right) + b_{i+1[j]}\\
    \hat{u}_{i+1[j]} &= \sum_k \left( W_{i+1[j, k]}^+ u^+_{i[k]} + W_{i+1[j,k]}^- l^+_{i[k]} \right) + b_{i+1[j]}
  \end{align}
  \label{eq:interval-anal}%
\end{subequations}
with the notation \mbox{$a^+=\max(a, 0)$} and \mbox{$a^-=\min(a, 0)$}.
Propagating the bounds through a ReLU activation is simply equivalent to
applying the ReLU to the bounds.

\paragraph{Planet Linear approximation}
An alternative way to obtain bounds is to use the relaxation of Planet. This is
the methods that was employed in the paper: we build incrementally the network
approximation, layer by layer. To obtain the bounds over an activation, we
optimize its value subject to the constraints of the relaxation.

Given that this is a convex problem, we will achieve the optimum. Given that it
is a relaxation, the optimum will be a valid bound for the activation (given
that the feasible domain of the relaxation includes the feasible domains subject
to the original constraints).

Once this value is obtained, we can use it to build the relaxation for the
following layers. We can build the linear approximation for the whole network
and extract the bounds for each activation to use in the encoding of the MIP.
While obtaining the bounds in this manner is more expensive than simply doing
interval analysis, the obtained bounds are better.

\subsection{Objective function}
In the paper, we have formalised the verification problem as a satisfiability
problem, equating the existence of a counterexample with the feasibility of the
output of a (potentially modified) network being negative.

In practice, it is beneficial to not simply formulate it as a feasibility
problem but as an optimization problem where the output of the network is
explicitly minimized.

\subsection{Comparison}

We present here a comparison on CollisionDetection and ACAS of the different
variants.

\begin{enumerate}
  \item \textbf{Planet-feasible} uses the encoding of
    Equation~\eqref{eq:mip-form}, with bounds obtained based on the planet
    relaxation, and solve the problem simply as a satisfiability problem.
  \item \textbf{Interval} is the same as \textbf{Planet-feasible}, except that
    the bounds used are obtained by interval analysis rather than with the
    Planet relaxation.
  \item \textbf{Planet-symfeasible} is the same as \textbf{Planet-feasible},
    except that the encoding is the one of Equation~\eqref{eq:altmip-form}.
  \item \textbf{Planet-opt} is the same as \textbf{Planet-feasible}, except that
    the problem is solved as an optimization problem. The MIP solver attempt to
    find the global minimum of the output of the network. Using Gurobi's
    callback, if a feasible solution is found with a negative value, the
    optimization is interrupted and the current solution is returned. This
    corresponds to the version that is reported in the main paper.
\end{enumerate}

\begin{figure}[h]
  \hfill%
  \begin{subfigure}[t]{.4 \textwidth}
    \includegraphics[width=\textwidth]{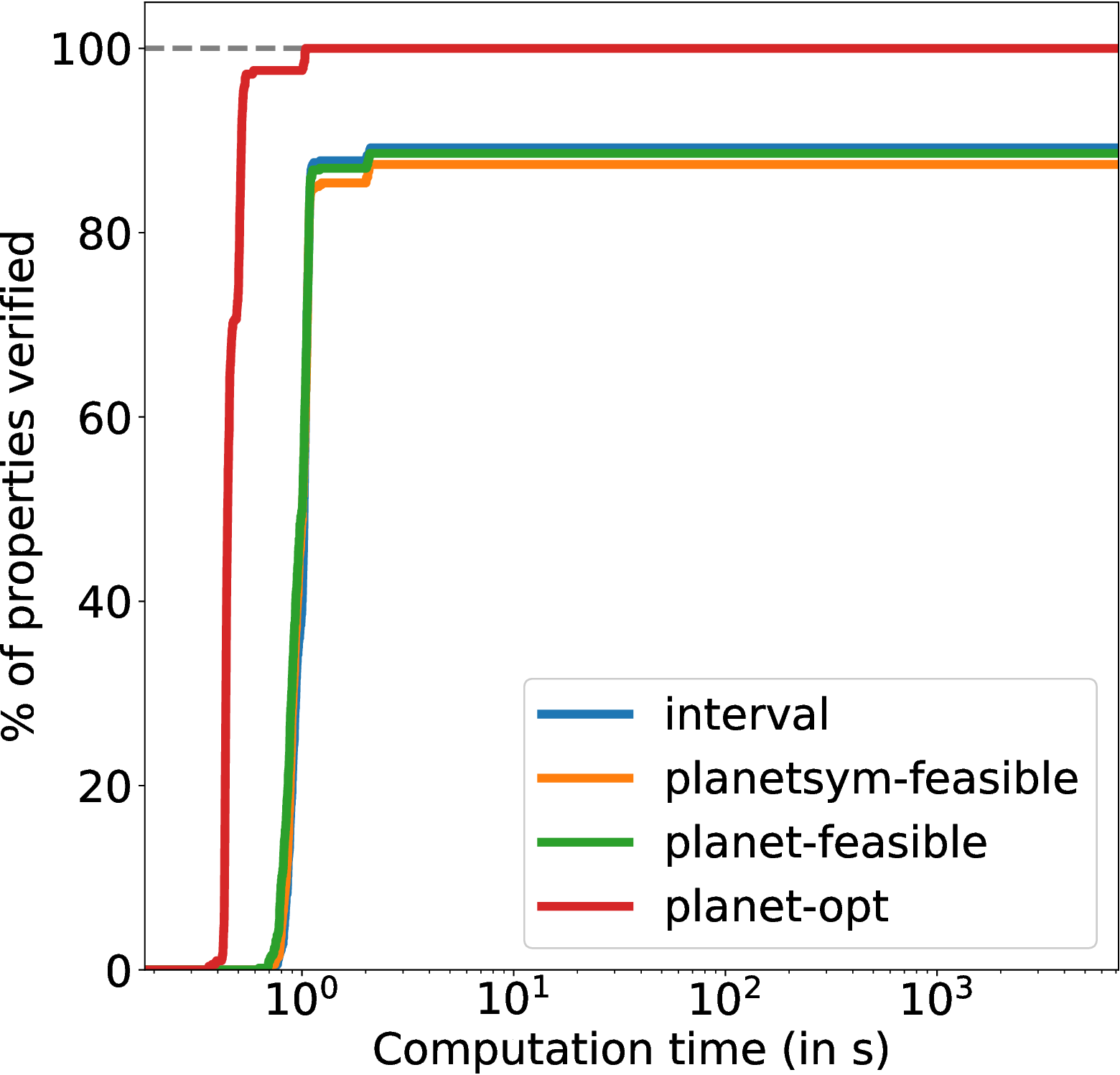}
    \caption{\label{fig:mip-collision-comparison} CollisionDetection Dataset}
  \end{subfigure}%
  \hfill%
  \begin{subfigure}[t]{.4 \textwidth}
    \includegraphics[width=\textwidth]{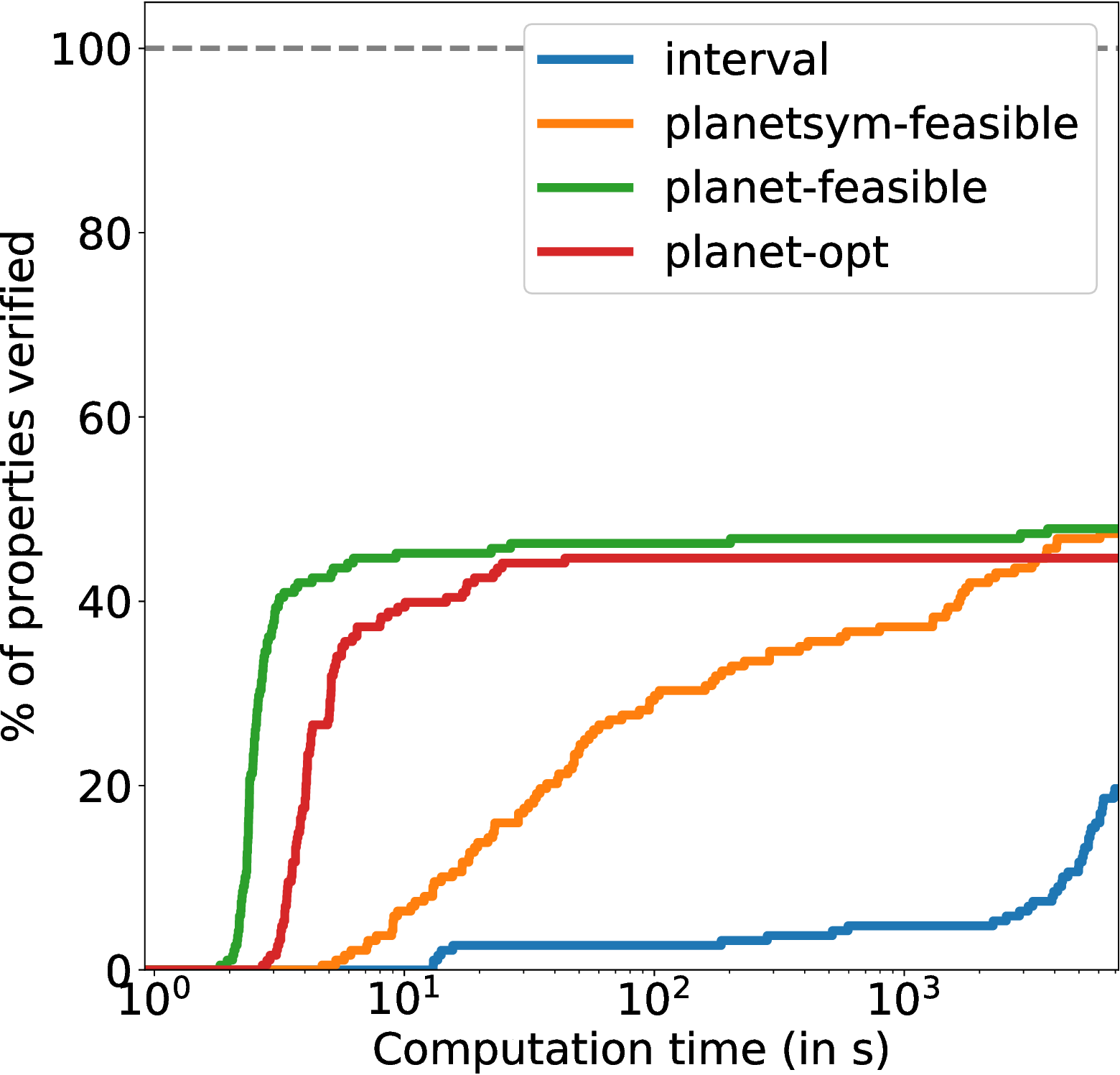}
    \caption{\label{fig:mip-acas-comparison} ACAS Dataset}
  \end{subfigure}%
  \hfill
  \caption{\label{fig:mip-comparison}\emph{\small Comparison between the different
      variants of MIP formulation for Neural Network verification.}}
\end{figure}

The first observation that can be made is that when we look at the
CollisionDetection dataset in Figure~\ref{fig:mip-collision-comparison}, only
\textbf{Planet-opt} solves the dataset to 100\% accuracy. The reason why the
other methods don't reach it is not because of timeout but because they return
spurious counterexamples. As they encode only satisfiability problem, they
terminate as soon as they identify a solution with a value of zero. Due to the
large constants involved in the big-M, those solutions are sometimes not
actually valid counterexamples. This is the same reason why the
\textbf{BlackBox} solver doesn't reach 100\% accuracy in the experiments
reported in the main paper.

The other results that we can observe is the impact of the quality of the bounds
when the networks get deeper, and the problem becomes therefore more complex,
such as in the ACAS dataset. \textbf{Interval} has the worst bounds and is much
slower than the other methods. \textbf{Planetsym-feasible}, with its slightly
worse bounds, performs worse than \textbf{Planet-feasible} and \textbf{Planet-opt}.


\section{Experimental setup details}
We provide in Table~\ref{tab:problem_size} the characteristics of all of the
datasets used for the experimental comparison in the main paper.
{
\begin{table}[h]
  \begin{minipage}[t]{.95\textwidth}
    \vskip 0pt
  \resizebox{\textwidth}{!}{
  \begin{tabular}{|c|c|c|}
    \hline
    \textbf{Data set} & \textbf{Count} & \textbf{Model Architecture}\\
    \hline
    \begin{tabular}{l}
      Collision\\
      Detection
   \end{tabular}& 500 & \begin{tabular}{@{}c@{}}6 inputs\\
                                 40 hidden unit layer, MaxPool \\
                                 19 hidden unit layer\\
                                 2 outputs\\
                               \end{tabular}\\
    \hline
    ACAS & 188 & \begin{tabular}{@{}c@{}}5 inputs\\
                   6 layers of 50 hidden units\\
                   5 outputs
                 \end{tabular} \\
    \hline
    PCAMNIST & 27 & \begin{tabular}{@{}c@{}} 10 or \{5, 10, 25, 100, 500, 784\} inputs\\
                      4 or \{2, 3, 4, 5, 6, 7\} layers \\
                      of 25 or \{10, 15, 25, 50, 100\} hidden units,\\
                      1 output, with a margin of +1000 or \\
                      \{-1e4, -1000, -100, -50, -10, -1 ,1, 10, 50, 100, 1000, 1e4\}\\
                      \end{tabular}\\
    \hline
  \end{tabular}}
\end{minipage}%
\hfill%
\begin{minipage}[t]{.95\textwidth}
  \caption{\label{tab:problem_size}\emph{Dimensions of all the data sets. For
    PCAMNIST, we use a base network with 10 inputs, 4 layers of 25 hidden units
    and a margin of 1000. We generate new problems by changing one
    parameter at a time, using the values inside the brackets.}}
\end{minipage}
\end{table}
}

\FloatBarrier

\section{Additional performance details}
Given that there is a significant difference in the way verification works for
SAT problems vs. UNSAT problems, we report also comparison results on the subset
of data sorted by decision type.

\begin{figure}[h]
  \begin{minipage}[b]{\textwidth}
    \vskip 0pt
    \centering
    \begin{subfigure}[t]{.50\textwidth}
      \vskip 0pt
      \includegraphics[width=.90\textwidth]{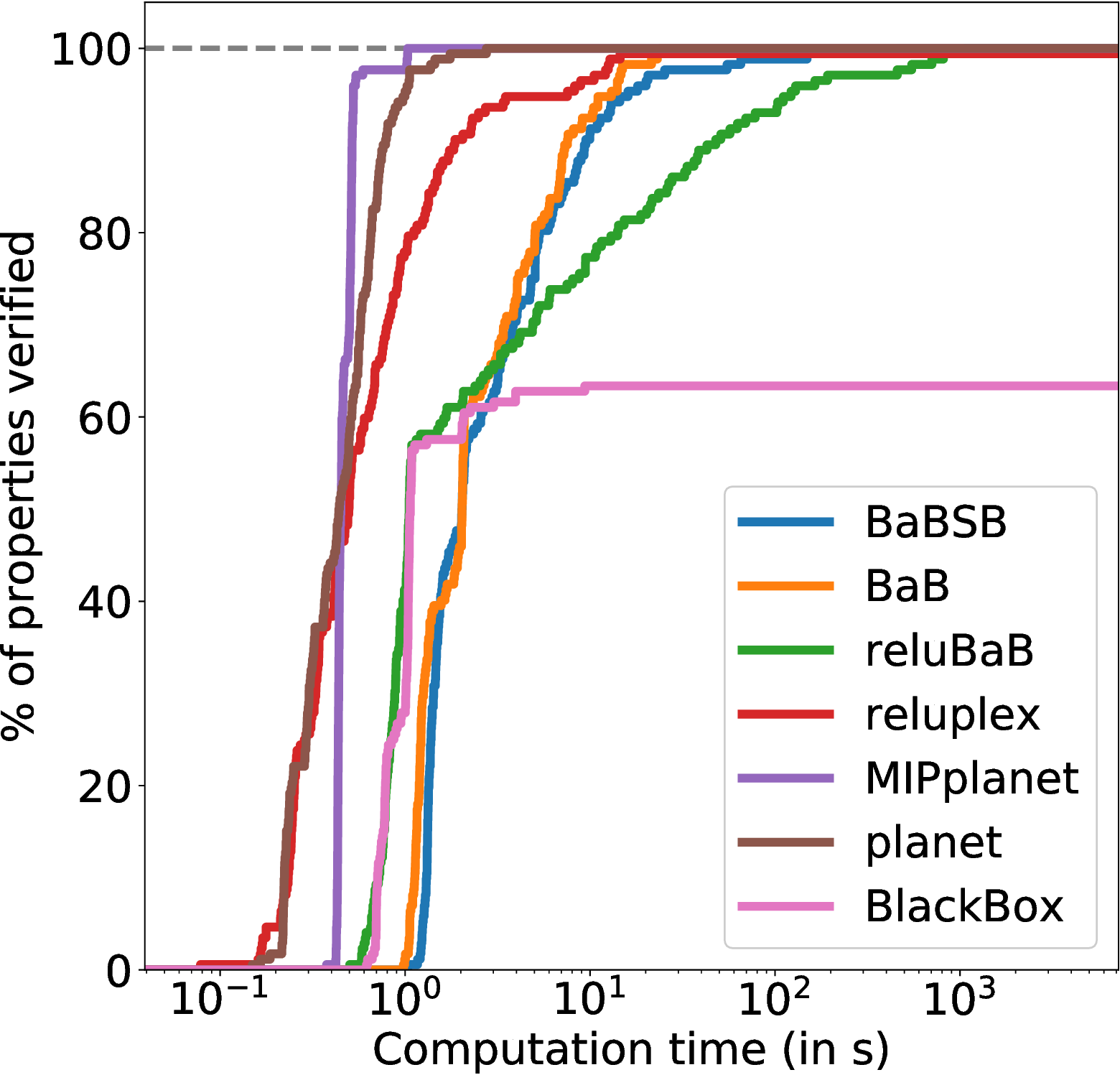}
      \caption{\label{fig:cactus_collision}\emph{On SAT properties}}
    \end{subfigure}%
    \begin{subfigure}[t]{.50\textwidth}
      \vskip 0pt
      \includegraphics[width=.90\textwidth]{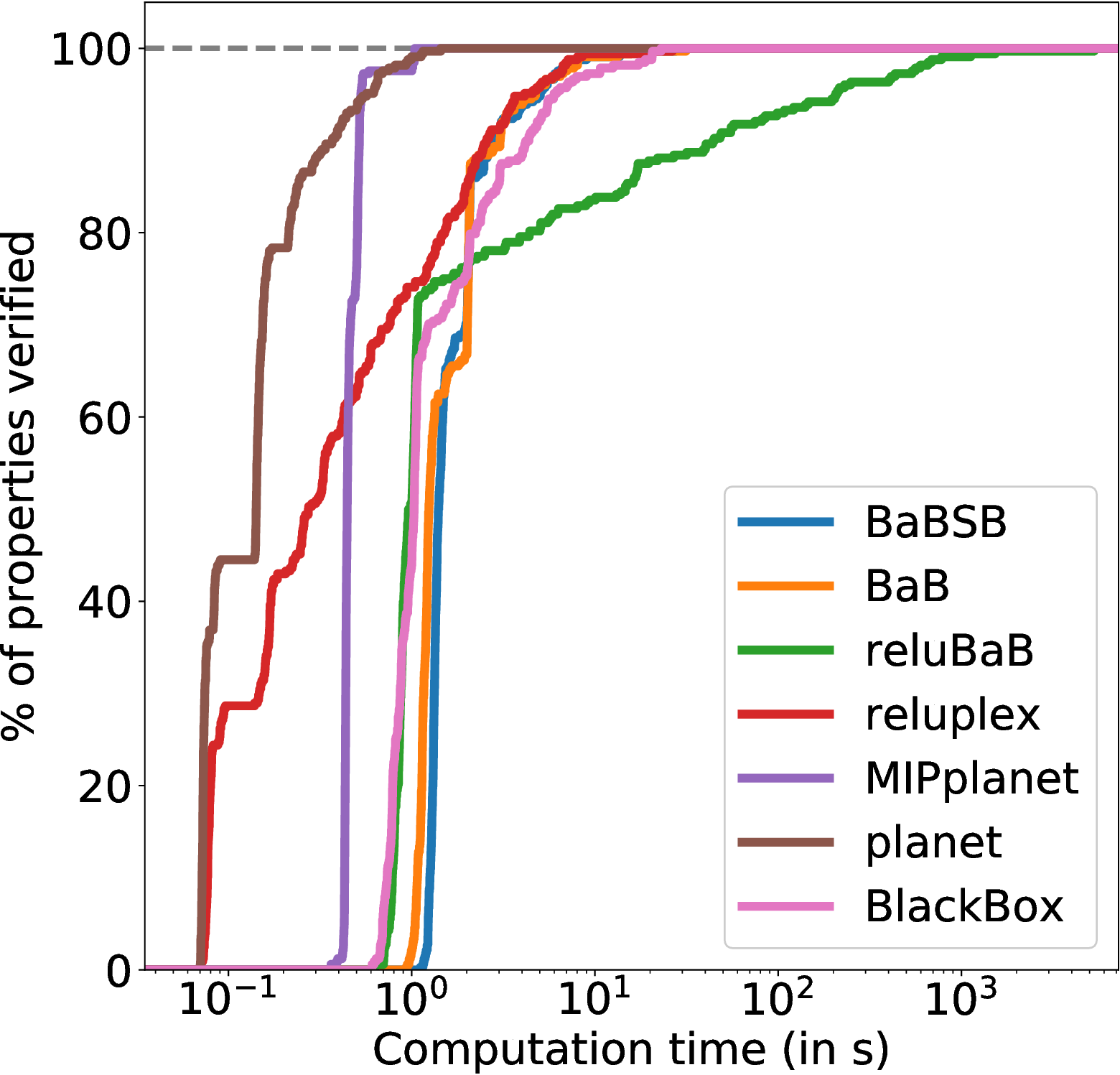}
      \caption{\label{fig:cactus_acas}\emph{On UNSAT properties}}
    \end{subfigure}
  \end{minipage}

  \begin{minipage}[t]{\textwidth}
    \vskip 0pt
    \caption{\label{fig:cactus_collision}\emph{ Proportion of properties
        verifiable for varying time budgets depending on the methods employed on
        the CollisionDetection dataset. We can identify that all the errors that
        \textbf{BlackBox} makes are on SAT properties, as it returns incorrect counterexamples.}}
  \end{minipage}
\end{figure}

\begin{figure}[h]
  \begin{minipage}[b]{\textwidth}
    \vskip 0pt
    \centering
    \begin{subfigure}[t]{.50\textwidth}
      \vskip 0pt
      \includegraphics[width=.90\textwidth]{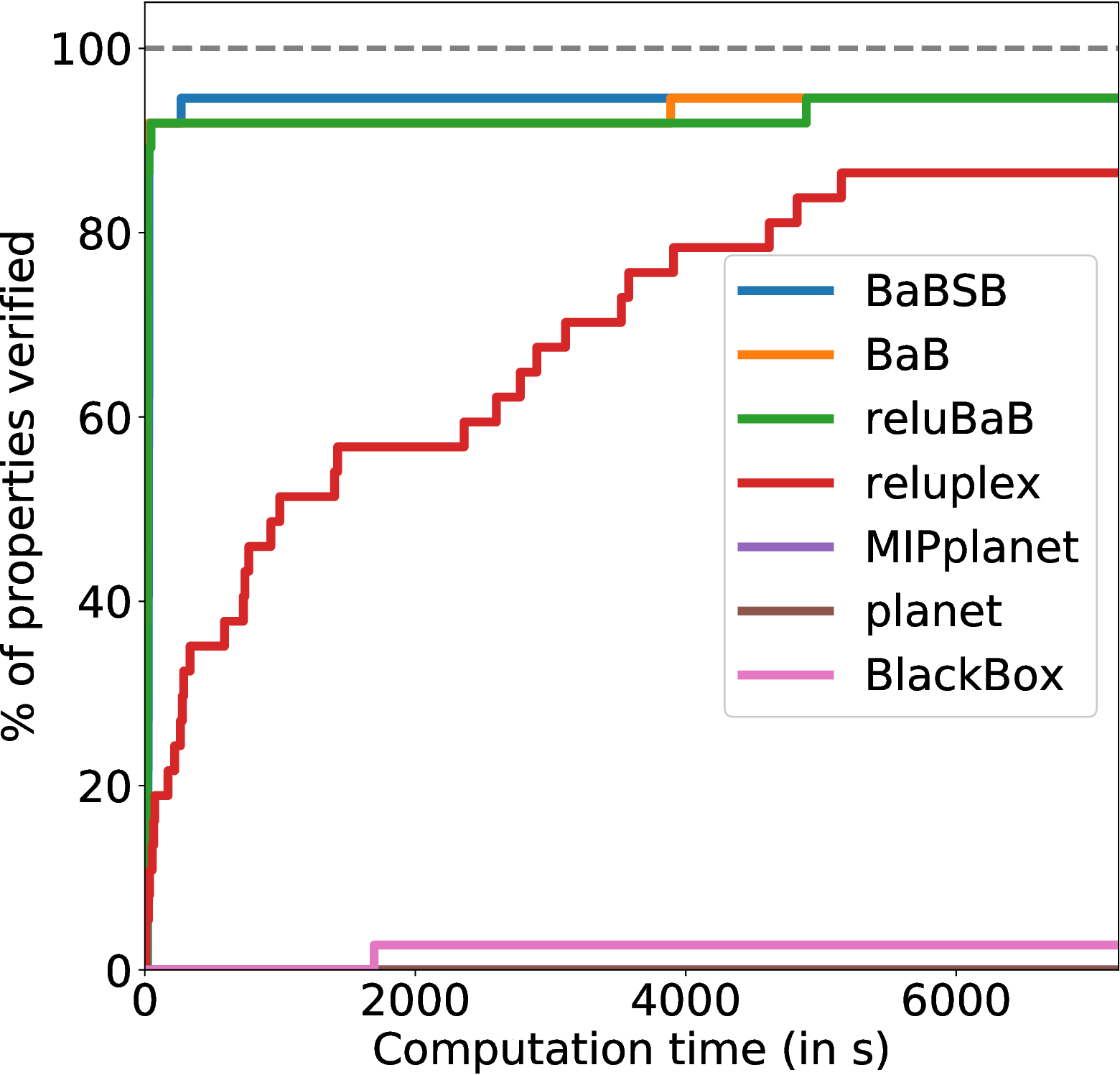}
      \caption{\label{fig:cactus_collision}\emph{On SAT properties}}
    \end{subfigure}%
    \begin{subfigure}[t]{.50\textwidth}
      \vskip 0pt
      \includegraphics[width=.90\textwidth]{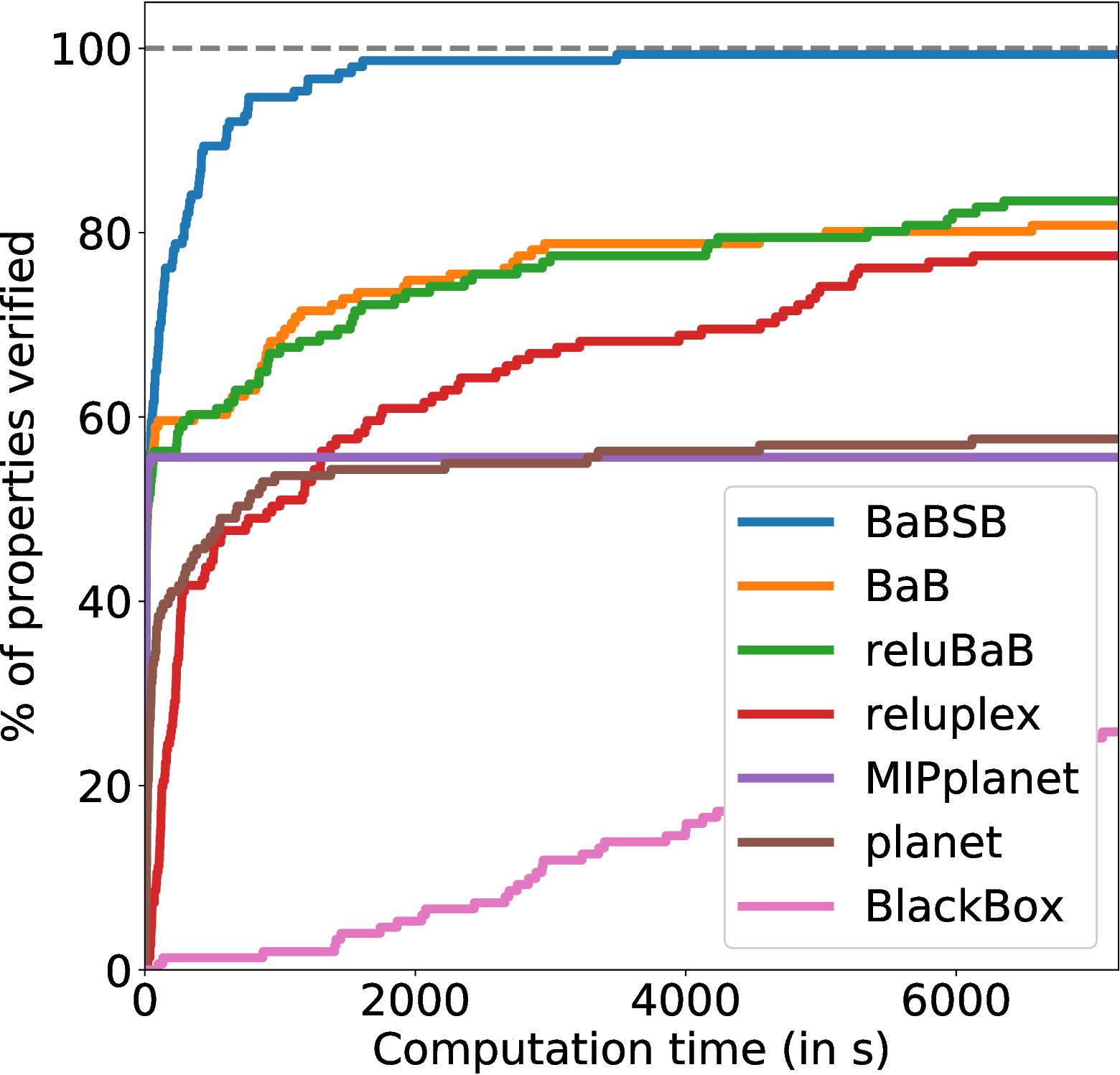}
      \caption{\label{fig:cactus_acas}\emph{On UNSAT properties}}
    \end{subfigure}
  \end{minipage}

  \begin{minipage}[t]{\textwidth}
    \vskip 0pt
    \caption{\label{fig:cactus_collision}\emph{ Proportion of properties
        verifiable for varying time budgets depending on the methods employed on
        the ACAS dataset. We observe that \textbf{planet} doesn't succeed in
        solving any of the SAT properties, while our proposed methods are
        extremely efficient at it, even if there remains some properties that
        they can't solve.}}
  \end{minipage}
\end{figure}

\section{PCAMNIST details}

\textbf{PCAMNIST} is a novel data set that we introduce to get a better
understanding of what factors influence the performance of various methods. It
is generated in a way to give control over different architecture parameters.
The networks takes $k$ features as input, corresponding to the first $k$
eigenvectors of a Principal Component Analysis decomposition of the digits from
the MNIST data set. We also vary the depth (number of layers), width (number of
hidden unit in each layer) of the networks. We train a different network for
each combination of parameters on the task of predicting the parity of the
presented digit. This results in the accuracies reported in
Table~\ref{tab:PCAMNIST-acc}.

The properties that we attempt to verify are whether there exists an input for
which the score assigned to the odd class is greater than the score of the even
class plus a large confidence. By tweaking the value of the confidence in the
properties, we can make the property either True or False, and we can choose by
how much is it true. This gives us the possibility of tweaking the ``margin'',
which represent a good measure of difficulty of a network.

\begin{figure}
  \centering
  \includegraphics[width=.5\textwidth]{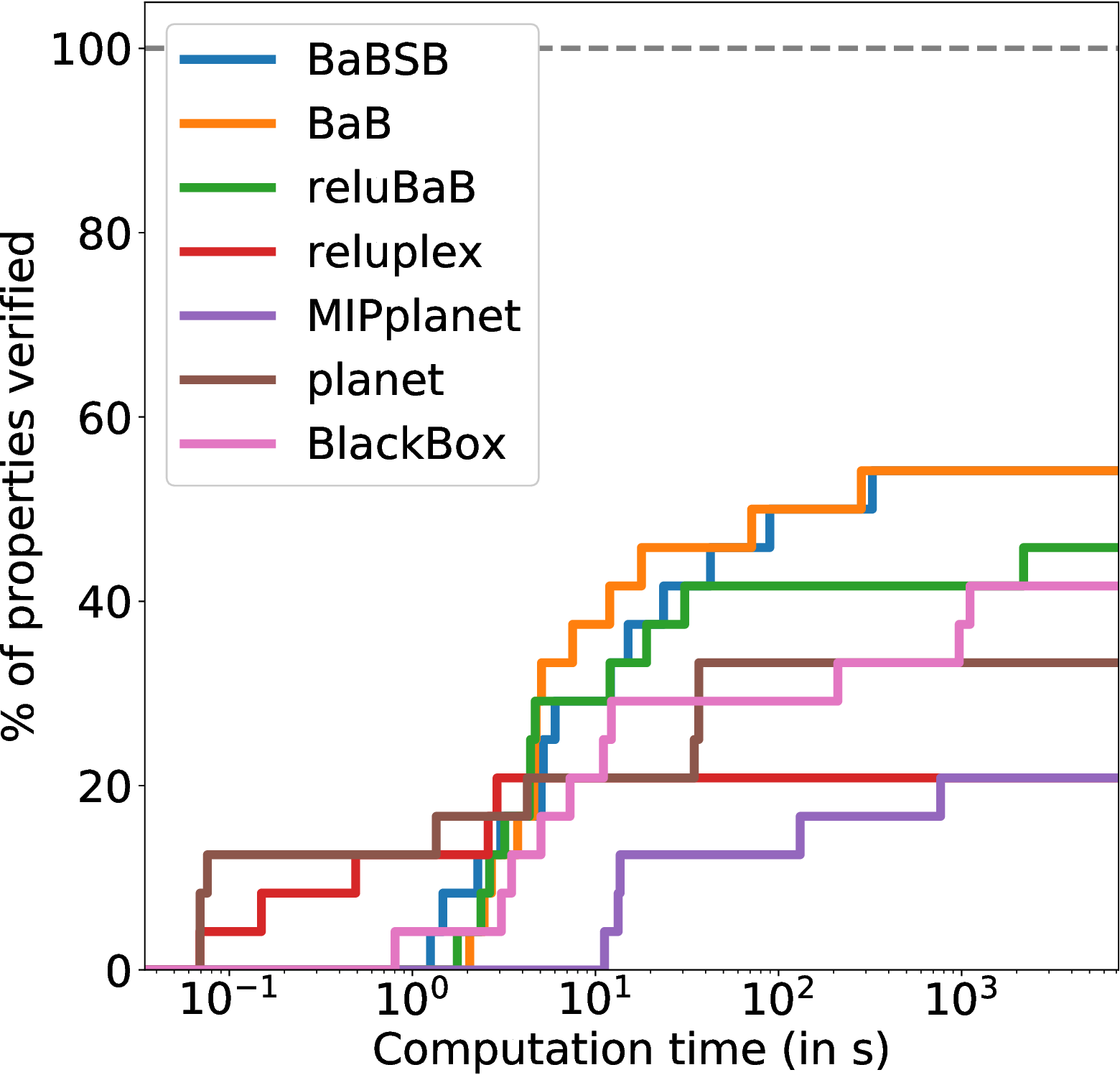}
  \caption{\label{fig:cactus_pcamnist}\emph{Proportion of properties verifiable
      for varying time budgets depending on the methods employed. The PCAMNIST
      dataset is challenging as None of the methods reaches more than 50\%
      success rate.}}
\end{figure}

In addition to the impact of each factors separately as was shown in the main
paper, we can also look at it as a generic dataset and plot the cactus plots
like for the other datasets. This can be found in Figure~\ref{fig:cactus_pcamnist}

\label{app:pcamnist-acc}
\begin{table}[h]
  \center
  \resizebox{.48\textwidth}{!}{
    \begin{tabular}{ccc@{\hspace*{4ex}}|cc}
      \toprule
      \multicolumn{3}{c|}{\textbf{Network Parameter}} & \multicolumn{2}{c}{\textbf{Accuracy}}\\
      \textbf{Nb inputs} & \textbf{Width} & \textbf{Depth} & \textbf{Train} & \textbf{Test} \\
      \midrule
      5 & 25 & 4 & 88.18\% & 87.3\%\\
      10 & 25 & 4 & 97.42\% & 96.09\%\\
      25 & 25 & 4 & 99.87\% & 98.69\%\\
      100 & 25 & 4 & 100\% & 98.77\%\\
      500 & 25 & 4 & 100\% & 98.84\%\\
      784 & 25 & 4 & 100\% & 98.64\%\\
      \midrule
      10 & 10 & 4 & 96.34\% & 95.75\%\\
      10 & 15 & 4 & 96.31\% & 95.81\%\\
      10 & 25 & 4 & 97.42\% & 96.09\%\\
      10 & 50 & 4 & 97.35\% & 96.0\%\\
      10 & 100 & 4 & 97.72\% & 95.75\%\\
      \midrule
      10 & 25 & 2 & 96.45\% & 95.71\%\\
      10 & 25 & 3 & 96.98\% & 96.05\%\\
      10 & 25 & 4 & 97.42\% & 96.09\%\\
      10 & 25 & 5 & 96.78\% & 95.9\%\\
      10 & 25 & 6 & 95.48\% & 95.2\%\\
      10 & 25 & 7 & 96.81\% & 96.07\%\\
      \bottomrule
    \end{tabular}}
  \caption{\label{tab:PCAMNIST-acc}Accuracies of the network trained for the
    \mbox{PCAMNIST} dataset.}
\end{table}

\end{document}